\documentclass[journal]{IEEEtran}

\usepackage{soul,framed} 

\usepackage[pdftex]{graphicx}
\graphicspath{{../pdf/}{../jpeg/}}
\DeclareGraphicsExtensions{.pdf,.jpeg,.png}

\usepackage[cmex10]{amsmath}
\usepackage{array}
\usepackage{mdwmath}
\usepackage{mdwtab}
\usepackage{eqparbox}
\usepackage{url}
\usepackage[]{algorithm2e}
\usepackage{algpseudocode}
\usepackage{listings}
\usepackage{graphicx} 

\usepackage{adjustbox}
\usepackage{graphicx}
\usepackage{caption}
\usepackage{subcaption}
 \usepackage[dvipsnames]{xcolor}
\usepackage[normalem]{ulem}
\useunder{\uline}{\ul}{}

\begin{document}
\bstctlcite{IEEEexample:BSTcontrol}
    \title{Multi-channel Emotion Analysis for Consensus Reaching in Group Movie Recommendation Systems}
  \author{Adilet~Yerkin,
      Elnara~Kadyrgali,
      Yerdauit~Torekhan,
      and ~Pakizar~Shamoi,~\IEEEmembership{Member,~IEEE,}

  \thanks{Manuscript received April 22, 2024. This paper is an expanded paper from the IEEE Digital Generation Student Conference held on April 4-5, 2024, in Astana, Kazakhstan.\textit{(Corresponding author: Pakizar Shamoi.)} }
  \thanks{A. Yerkin, E. Kadyrgali, Y. Torekhan, and P. Shamoi are with the School of Information Technology and Engineering, Kazakh-British Technical University, Almaty 050000, Kazakhstan (e-mail: p.shamoi@kbtu.kz).}}

\markboth{IEEE TRANSACTIONS ON COMPUTATIONAL SOCIAL SYSTEMS, VOL.~0, NO.~0, May~2024
}{Yerkin \MakeLowercase{\textit{et al.}}: Multi-channel Emotion Analysis for Consensus
Reaching in Group Movie Recommendation
Systems}

\maketitle

\begin{abstract}
Watching movies is one of the social activities typically done in groups. Emotion is the most vital factor that affects movie viewers' preferences. So, the emotional aspect of the movie needs to be determined and analyzed for further recommendations. It can be challenging to choose a movie that appeals to the emotions of a diverse group. Reaching an agreement for a group can be difficult due to the various genres and choices. This paper proposes a novel approach to group movie suggestions by examining emotions from three different channels: movie descriptions (text), soundtracks (audio), and posters (image). We employ the Jaccard similarity index to match each participant's emotional preferences to prospective movie choices, followed by a fuzzy inference technique to determine group consensus. We use a weighted integration process for the fusion of emotion scores from diverse data types. Then, group movie recommendation is based on prevailing emotions and viewers' best-loved movies. After determining the recommendations, the group's consensus level is calculated using a fuzzy inference system, taking participants' feedback as input. Participants (n=130) in the survey were provided with different emotion categories and asked to select the emotions best suited for particular movies (n=12). Comparison results between predicted and actual scores demonstrate the efficiency of using emotion detection for this problem (Jaccard similarity index = 0.76). We explored the relationship between induced emotions and movie popularity as an additional experiment, analyzing emotion distribution in 100 popular movies from the TMDB database. Such systems can potentially improve the accuracy of movie recommendation systems and achieve a high level of consensus among participants with diverse preferences.
\end{abstract}

\IEEEpeerreviewmaketitle
\section{Introduction}
\IEEEPARstart{I}{n} today's world, movies have a significant impact on cultural conversations and personal enjoyment. With plenty of options available, individuals choose what to watch, guided by their tastes and the influences around them. Their current context influences the user's decision-making process, but using the internet helps to simplify this task.
At the same time, an increasing number of artificial intelligence (AI) systems are expected to depend on identifying emotions via speech, facial expressions, and conversation content \cite{ieee1}. 

Emotions play a key role in successful information delivery, communication, and engagement and can critically affect the perception and attitudes of viewers \cite{perc}. Researchers are increasingly interested in multimodal emotion identification because of its potential to overcome the limitations of monomodal systems \cite{perc2}, \cite{perc3}. Especially visual emotion analysis has recently received much attention, thanks to the growing trend of expressing and understanding emotions through photographs on social networks \cite{perc3}.

Emotional data is crucial in many multimedia applications \cite{ieeecs4}. Emotion is the most important element that links movies and humans. In today's entertainment industry, watching a movie is not only about experiencing audio and visual sensations; it explores the complex domain of human emotions. Watching television shows or movies is an example of a social activity usually done in groups \cite{3}. The process of selecting a movie that satisfies the emotional preferences of a diverse group can be challenging. Facilitating agreement among group members is a significant challenge in recommending movies. Emotions play a crucial role since we primarily watch movies to experience emotions.

Nowadays, recommendation systems are omnipresent due to the excessive volume of multimedia information. When selecting a movie, reaching a consensus among people with different preferences can be challenging.  Previous studies revealed that the decision-making process becomes more difficult in the case of uncertain product features (e.g., delivered emotions). For this reason, when choosing which media content to interact with, users tend to rely on the insights provided by collaborative input \cite{metzger2010social}.




This paper presents a novel methodology for group movie recommendations. We analyze emotions from three channels: movie descriptions, soundtracks, and posters. We use the Jaccard similarity index to match each participant's emotional preferences with potential movie choices and then apply a fuzzy inference system to determine group consensus. In our previous works, we introduced the approach for movie emotion recognition \cite{arxiv1}. The main contributions of this work are:



\begin{enumerate}
\setcounter{enumi}{0}
\item Proposing a novel approach for achieving consensus in group movie selection using a multi-channel emotion recognition approach based on movie data like poster, text description, and soundtrack.
\item \textit{Consensus Estimation}. After determining the most recommended movie for the group, the level of consensus is calculated considering participants' feedback, namely, agreement and confidence levels. We propose a fuzzy inference system for this purpose.
\item \textit{Emotion-Popularity analysis in movies.} We also investigate which emotions conveyed through poster colors (color image features), the soundtrack (music features), or the movie description (text features)—have the greatest impact on movie popularity.

\end{enumerate}

The paper is organized as follows. The present
section is an introduction that contains the research goals and main contributions. Section II involves a comprehensive literature review on emotion analysis, group movie recommendations, and consensus building. Section III describes the methods and details of the data collection. The experimental investigation, examples, and results are presented in Section IV. Section V comprises discussions on the topic, and Section VI includes a conclusion and ideas for future development.

\section{Related Work}


\subsection{Emotion Analysis}
Studies on emotion recognition can be categorized based on the approaches used in research. The first category of studies mainly focuses on analyzing a single component to detect the emotion. On the other hand, the second category employs multi-componential approaches in emotion analysis, integrating more human-like applications.

Several movie analysis-related works employed emotion recognition for various purposes. One study presents a multiclass emotion classifier emphasizing negative emotions, using a rich set of metadata and content from a tagged movie transcript \cite{ieee1}. Another study introduces a bimodal method that combines facial expression recognition (FER) with the actors' "semantic orientation" in their dialogue to recognize emotions in scenarios with challenging lighting, position changes, and occlusions \cite{ieee2}. An emotion map for a movie based on aggregating and projecting reviewers' movie scores and reviews onto the movie was proposed in \cite{ieee3}. An approach for emotion tracking using supervised learning methods was introduced to model the continuous affective response using hidden Markov Models \cite{ieee4}. Multichannel modeling approaches for predicting movie-induced and perceived emotions were presented in \cite{ieee5}.  Some works classify emotions present in movies using EEG signal analysis \cite{eeg}, \cite{eeg2}. Another study introduces an EEG-based user-independent emotion recognition method to recover emotion tags for videos, pupillary response, and gaze distance. The study employed 20 movie video clips and 24 human subjects watching these emotional video clips. Ground truth was defined based on the median arousal and valence scores to define the ground truth from the survey. The other study proposes a multi-label positive emotion classification method based on the brain activities of viewers exposed to affective content from movies, TransEEG, a model for multi-label positive emotion classification from a viewer's brain activities when watching emotional movies \cite{ieeecs4}.

Considering the influence of multimedia information on the audience's emotions, the new methodology was proposed for movie preference prediction using low-level multimedia feature extraction and SVM-based classification and applied to a collection of 725 movie trailers \cite{ieeecs1}. The following features were used: trailer colors, motion, and shot.
\subsection{Group Movie Recommendation}
Group Recommendation systems (RS) are RSs that incorporate group preferences, particularly in the context of movies. RSs are often divided into two types: collaborative filtering (CF) and content-based filtering (CBF). Some of the constraints of these traditional systems include the requirement for previous user information to complete the recommendation assignment \cite{sentiment}. 


Several studies have examined the problem of group movie recommendation using various methods.

The study \cite{1} presents "Happy movie," a Facebook-integrated app that uses personality, social trust, and past preferences to enhance group movie recommendations. The goal is to improve consensus and overcome the limitations of existing systems. One study proposes using movie posters to improve movie recommender systems, addressing issues of information overload, sparsity, and cold-start problems \cite{2}. The study introduces a recommendation system for moviegoers in temporary groups, using the Slope One algorithm for individual predictions and the Multiplicative Utilitarian Strategy for group recommendations \cite{3}.

Some studies propose the use of DL, ML, and NLP techniques for this task \cite{6}, \cite{7}, \cite{new20}. A novel method for suggesting movies involves using a knowledge graph that captures human emotions from movie reviews \cite{6}. This is accomplished through machine learning techniques. A chatbot prototype can provide tailored movie recommendations by incorporating the emotional states of users, which are extracted from chat messages, with the knowledge graph. A study by \cite{7} presented a model that uses social networks, microblogging data, and sentiment analysis to improve program recommendations on media sites such as YouTube and Hulu. This model addresses the "cold-start" problem by analyzing user preferences for similarity between movies and TV episodes.


Another study \cite{GAN2021114695} introduced the Dynamic Interest Flow (DIF) system, a novel approach for adapting movie recommendations to users' evolving preferences by leveraging the UMIS (User Movie Interest Space). It captures the intrinsic characteristics and relationships of interests that influence users' movie-watching decisions. It facilitates tracking users' interest evolution alongside their rating histories to more accurately forecast future rating events. An actor-based recommendation system was proposed using content-based filtering that considers the genres of 509 South Korean movies and the movieography information of the actor \cite{ieeecs2}. 

Multiple studies focused on genre classification as a basis for proving movie recommendations for groups \cite{4,5}. Studies related to emotions reveal that mood states notably influence movie ratings, with variations observed across genres \cite{WINOTO20106086}; for instance, positive moods tend to elevate ratings for romantic comedies, while tiredness may increase ratings for crime thrillers. Incorporating mood states suggested by the PANAS-X inventory into a mood-aware recommendation mechanism has been shown to enhance recommendation accuracy by accounting for these mood-related biases.

A hybrid approach to movie recommendations was proposed \cite{new21}, integrating tags and human ratings to address the limitations of current services that often overlook the depth of user annotations \cite{8}. There are several methods to improve recommendation systems. Some of these methods include combining k-means clustering and genetic algorithms \cite{9}, user-based collaborative filtering using both ratings and social connections to calculate user similarity \cite{10}, using graph attention network (GAT) \cite{19}, integrating basic movie attributes such as genre, cast, director, keywords, and descriptions along with ratings using a CNN-based deep learning model \cite{12}, and using an item-based collaborative filtering approach \cite{new19}.

Another representative of a hybrid recommendation system for movies integrates collaborative filtering, content-based filtering, and sentiment analysis of tweets from microblogging platforms \cite{sentiment}. The idea is to capture current trends, public sentiment, and audience reactions to movies.

\begin{figure*}[t!]
    \centering
    \includegraphics[width=\textwidth]{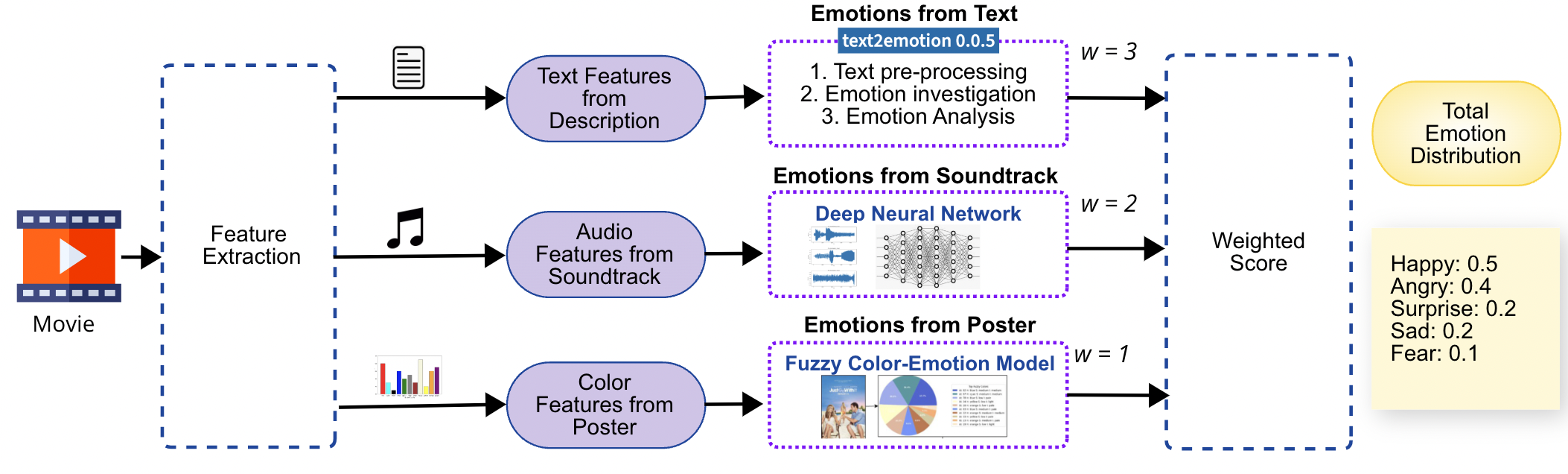}
    \caption{Multi-channel Emotion Recognition from movie data}
    \label{figure1}
\end{figure*}

Some studies focused on integrating consensus estimation in providing recommendations. Recent research presents a hybrid recommendation system that combines collaborative and content-based content, including top critic consensus and movie rating scores \cite{consensus1}.

It has been observed that there are only a few studies that concentrated on making collective suggestions for movies based on emotional features. To address this problem, our paper presents a new technique that uses emotional analysis to assist in the group movie selection process. By analyzing emotional data from multiple sources such as movie posters, main soundtrack, and descriptions, our methodology provides a comprehensive understanding of the emotional landscape associated with each movie.

Our approach is based on the understanding that movies are created to elicit specific emotional responses from the audience. movie posters capture the essence of the movie, soundtracks enhance the emotional tone, and descriptions provide the context. When combined, they create a complex emotional profile. Our methodology aims to integrate these elements, allowing us to match a movie's emotional tone with a group's collective mood and preferences. This simplifies the decision-making process regarding choosing a movie to watch.



\section{Methods}

Our framework consists of three steps:
\begin{itemize}
    \item Multi-channel Emotion Detection. Three emotion channels are used to identify the emotions in the movie: the poster's color features, the soundtrack's audio features, and the movie description's text features. As can be seen from the methodology presented in Fig. \ref{figure1}, all characteristics extracted from three tracks are aggregated using the weighted score. 
    \item Recommendation using Jaccard coefficient based on detected emotions. 
We also integrated genre as a normalizing factor in the decision step to make precise recommendations.

    \item Consensus evaluation using Fuzzy Inference System.
\end{itemize}






\subsection{Data Collection}
\subsubsection{IMDB}
We collected data from the Internet Movie Database (IMDb) \cite{imdb}, an online repository of information about movies, television shows, and other information associated with them. We selected 12 movies of different genres and years to evaluate the model, including \textit{Comedy, Romance, Horror, Drama,} and \textit{Fantasy}. The database provided three main elements for each movie: the plot summary, posters, and a 30-second excerpt from one of the soundtracks.

 Analyzing a movie's original soundtrack is not always possible, while the other two channels are always discoverable. So, for some movies, we chose music in trailers or during the movie itself instead of the original soundtrack. The following movies were selected for further analysis: Titanic (1997), Bride Wars (2009), Insidious: Chapter 3 (2015), Annabelle: Creation (2017), Just Go With It (2011), Me Before You (2016), Interstellar (2014), Edge of Tomorrow (2014), Passengers (2016), Don’t Breathe 2 (2021), The Proposal (2009), and The Holiday (2006). After considering various unofficial rate charts online, movies that garnered significant attention from global audiences between 1997 and 2021 were selected.

\begin{table}[tb]
\centering
\caption{Example of records on top 100 popular movies from TMDB}
\resizebox{\columnwidth}{!}{%
\begin{tabular}{|p{0.35cm}|p{1.7cm}|p{1.7cm}|p{2cm}|p{1.3cm}|p{2.3cm}|}
\hline
\textbf{id} & \textbf{Title}  & \textbf{Overview}      & \textbf{Genres}                                    & \textbf{track} & \textbf{Performer}                      \\ \hline
1           & Kung Fu Panda 4 & Po is gearing up to... & Action, Adventure, Animation, Comedy, Family       & Journey        & Hans Zimmer, Steve Mazzaro              \\ \hline
2           & Damsel          & A young woman's...     & Fantasy, Adventure, Action                         & Elodie's Maze  & David Fleming                           \\ \hline
3           & Sri Asih        & Alana discover the...  & Action, Adventure, Science Fiction, Fantasy, Drama & Yatim          & Aghi Narottama, Bemby Gusti, Tony Merle \\ \hline
4           & No Way Up       & Characters from...     & Action, Horror, Thriller                           & Opening Titles & Andy Gray                               \\ \hline
5           & Argylle         & When the plots of...   & Action, Adventure, Comedy                          & Moke Mayhem    & Lorne Balfe                             \\ \hline
... & ... & ... & ... & ... & ... \\ \hline
100         & My Fault        & Noah must leave her... & Drama, Romance                                     & Me quiero ir   & lusillón                                \\ \hline
\end{tabular}%
}
\label{pop-table}
\end{table}

\subsubsection{TMDB}

For our experiment on exploring the relationship between induced emotions and movie popularity, we also collected the data using community-built The Movie Database (TMDB)\footnote{https://www.themoviedb.org} with 904,555 movie metadata except for adult content and Deezer\footnote{https://www.deezer.com} music database. TMDB has a voting score from the community, therefore we used this data to analyze the correlation between the popularity of the movie and its emotional range and to represent the emotion distribution in 100 popular movies. Table \ref{pop-table} shows a snippet of the database containing 100 popular movies' details. The movie's overview as text, poster images in '.jpg' format, and a list of genres are taken from TMDB via the database's API. As TMDB does not provide the soundtrack data, a 30-second clip from the original soundtrack was taken from Deezer API in '.mp3' format, requested by the track's name and its performer, and transferred to '.wav'.

\subsection{Emotion Detection}

Human emotions are typically expressed and perceived through numerous modalities \cite{perc}.  Algorithm I provides the general pseudocode for emotion detection from movies.

\subsubsection{Emotions in movie description}
We utilized text2emotion version 0.0.5 \cite{t2e} to identify the emotions in the movie description. Five emotion categories were used: \textit{Happy, Angry, Sad, Surprise,} and \textit{Fear}. Text2emotions provides a dictionary with emotion categories as keys and scores for each category as values. Only emotions with non-zero scores are considered.

The movie overview's text emotions were identified using text2emotion 0.0.5 \cite{t2e}, a Python library that helps to extract the emotions from textual content. Taking as input data to process, it returns the dictionary with keys representing emotion categories and corresponding scores for each emotion category offered by text2emotion. The aggregation of the scores is always 1, so the percentage of each emotion in the input text can be observed. The five emotion categories were used: \textit{Happy, Angry, Sad, Surprised,} and \textit{Fearful}  Not every one of the five emotions is covered in the text, so only scores that are not 0 are considered.

\begin{algorithm}[tb]
\SetAlgoLined
\KwIn{Raw features for each movie (poster, audio, text) in dataset}
\KwOut{Dictionary with weighted scores for each emotion for each movie}

\BlankLine
\SetKwFunction{FindAudioEmotion}{FindAudioEmotion}
\SetKwFunction{FindImgEmotion}{FindImgEmotion}
\SetKwFunction{FindTextEmotion}{FindTextEmotion}

\BlankLine
\textbf{Initialization:}
Dictionary $emotionScores$ for storing emotion scores for each movie\;

\BlankLine
\While{there are more movies}{
    Get features for the current movie\;
    emotionAudio $\leftarrow$ \FindAudioEmotion ;\\
    emotionPoster $\leftarrow$ \FindImgEmotion ;\\
    emotionText $\leftarrow$ \FindTextEmotion 
    

    w $\leftarrow$ weighted average; \\
    \tcp{Here we find a weighted average for each emotion from 3 channels}

    Store emotion scores for the current movie in dictionary \emph{emotionScores}\;
}
\Return $emotionScores$;

\BlankLine
\SetKwProg{Fn}{Function}{:}{}
\Fn{\FindAudioEmotion}{
    \tcp{Deep Neural network that takes audio from a movie, slices it, and processes it}
}

\BlankLine
\SetKwProg{Fn}{Function}{:}{}
\Fn{\FindImgEmotion}{
    \tcp{Fuzzy Color Emotion Model that extracts a dominant color palette and correlates it with emotion}
}

\BlankLine
\SetKwProg{Fn}{Function}{:}{}
\Fn{\FindTextEmotion}{
    
    
    \tcp{Here we use text2emotion to analyze the emotions present in the text}
}

\label{alg}
\caption{Multi-channel emotion analysis of movies}
\end{algorithm}

\subsubsection{Emotions in poster}

\begin{figure*}[h]
    \centering
    \includegraphics[width=\textwidth]{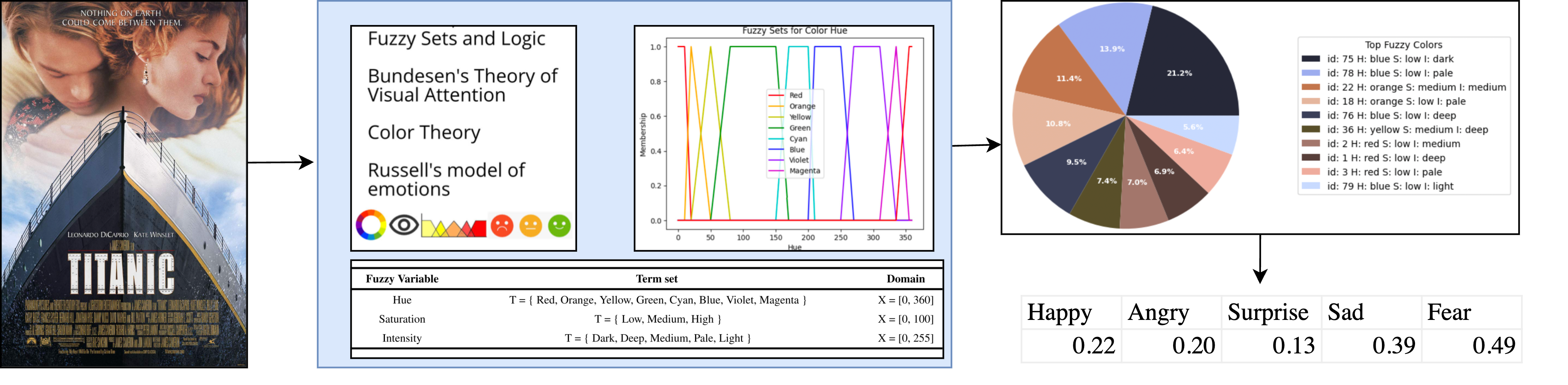}
    \caption{Example of emotion extraction from the image. Method is adapted from \cite{muratbekova2023color}}
    \label{figure2}
\end{figure*}

We used a novel fuzzy sets-based method to categorize emotions based on colors in a movie poster \cite{muratbekova2023color}, which fits well with human assessments' imprecise and subjective nature. The proposed approach can be easily adapted to suit our specific needs. The research employed fuzzy colors \cite{18}, \cite{shamoi2023universal} (with n=120) and a range of emotions (with n=10) to determine fuzzy color distributions for ten distinct emotions, namely \textit{Anger, Shyness, Happiness, Sadness, Gratitude, Shame, Fear, Trust, Love,} and \textit{Surprise}. After processing, they are transformed into a crisp domain, and we gain a knowledge base of primary color-emotion correlations. The study revealed strong correlations between specific emotions and colors (2AFC score = 0.77).
We use these correlations and Jaccard's similarity to identify emotions in poster images. To fit our context, we focus on a specific set of emotions listed in reference \cite{10}: \textit{Anger, Happiness, Sadness, Fear, Love,} and \textit{Surprise}. Our process is illustrated in Fig. \ref{figure2}.

\subsubsection{Emotions in movie soundtrack}

\begin{figure*}[h]
    \centering
    \includegraphics[width=\textwidth]{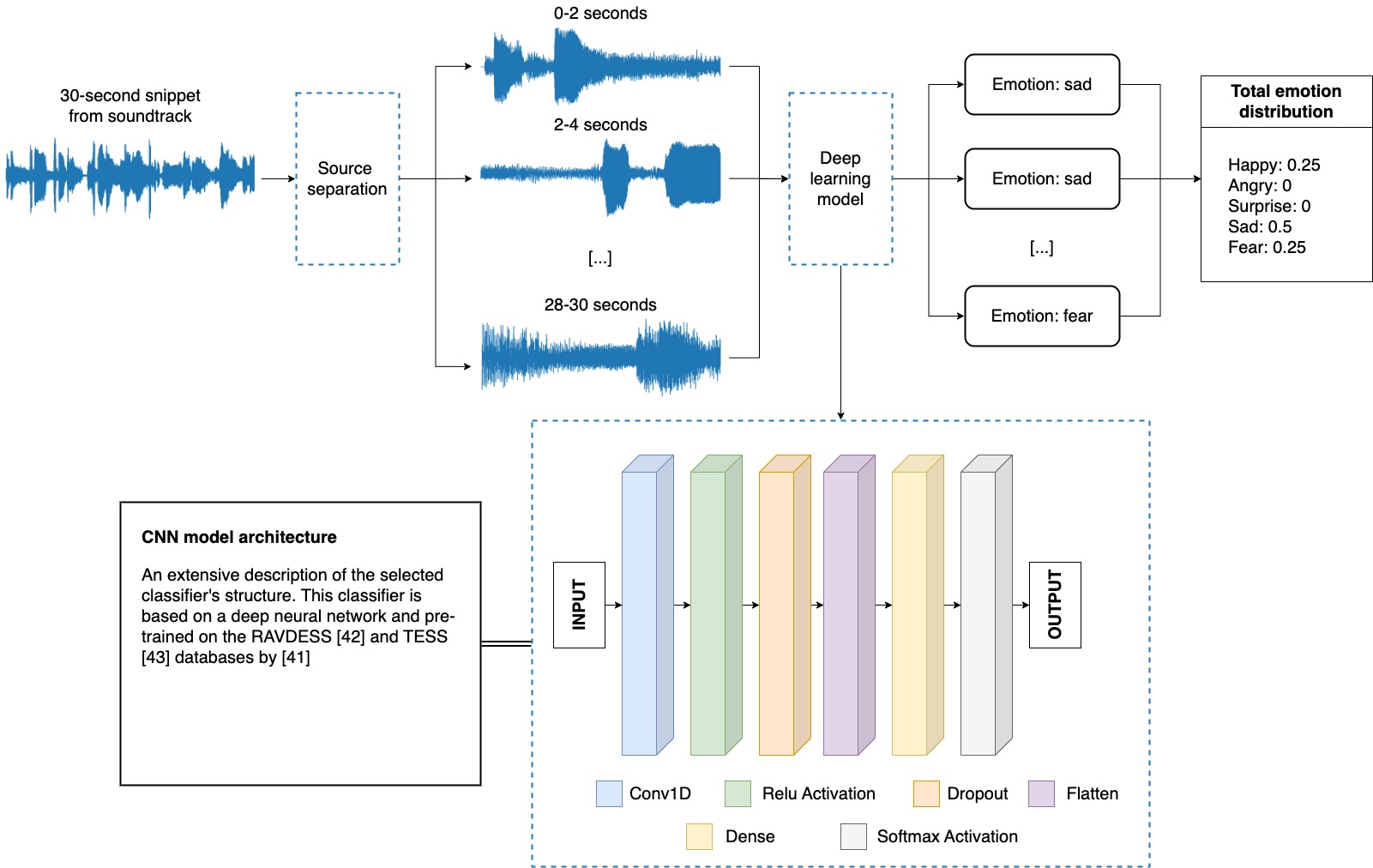}
    \caption{Audio Emotion Recognition using deep learning}
    \label{audio_emotion}
\end{figure*}

Several methods and strategies are used in Music  Emotion Recognition(MER). 
In this research, we conduct single-component research simply applying audio analysis to our soundtrack excerpts, preliminarily dividing them into 2-second partitions. The procedure for detecting emotions from audio recordings is shown in Fig. \ref{audio_emotion}.

Music emotion recognition employs low-level features such as beat, pitch, rhythm, valence, and tempo \cite{17}. Soundtrack analysis was performed using a pre-trained model based on a deep neural network algorithm, performed in \cite{16} that obtained an F1 score of 0.91 on the authors’ test set. The lowest performance was obtained in the sad emotional class with a score of 0.87, and the best results were seen in the angry class with a 0.95 score. The original paper \cite{16} uses only the RAVDESS dataset. In contrast, the current paper's model uses an updated one from GitHub, trained on two different datasets, with an F1 score of 0.80. So, the model is trained to recognize eight emotion categories (\textit{Neutral, Calm, Happy, Sad, Angry, Fearful, Disgust,} and \textit{Surprise}) using the Ryerson Audio-Visual Database of Emotional Speech and Song (RAVDESS) dataset \cite{dataset_train} and Toronto emotional speech set (TESS) \cite{SP2/E8H2MF_2020}. Important to mention that the model was initially intended for speech analysis, we noticed that it also works well for songs.

We divided the initial 30-second duration audio files into 15 partitions for accurate analysis. As seen from Fig. \ref{audio_emotion}, in a 30-second song, it is difficult to inspect features due to an enormous variation in the wave plot. This means that there can be several emotions in one partition of the whole soundtrack, so to hold as many emotions as possible,  we analyze each 2 seconds of the song separately. Many emotions in a single input can result in emotions such as \textit{Disgust}, and the other emotions will be lost in the output. The network may process vectors containing 40 audio features representing the audio frame's numerical form for every audio file supplied as input. We use the main five emotions (\textit{Happy, Angry, Sad, Surprise,} and \textit{Fear}) from \cite{16}, which is our main focus. The model generates an emotional class for input audio excerpts and encodes it as follows: \textit{Happy}=2; \textit{Sad}=3; \textit{Angry}=4; \textit{Fearful}=5; \textit{Surprised}=7. We have observed ten emotional labels in each of the 15 parts of a single excerpt. Based on the prevalence of emotional categories, we have formulated a dictionary with scores for each category, represented as keys similar to the output of text emotion analysis.

\subsubsection{Multimodal Fusion}
We use a weighted integration process to achieve a cohesive result for the fusion of emotion scores from diverse data types.


The emotional scores from the movie poster, main soundtrack, and movie description are combined using a weighted average approach \eqref{agg}. Let $E_{p}$, $E_{m}$, and $E_{d}$ denote the emotional scores from the poster, main soundtrack, and movie description. Each movie's final emotional score ($E_{agg}$) is calculated by adding the individual emotional scores multiplied by a particular weight.

\begin{equation} \label{agg}
E_{agg} =  \frac{w_{p}  *  E_{p}  +  w_{m}  *  E_{m} + w_{d}  *  E_{d}}{w_{p}  +  w_{m}  +  w_{d}}
\end{equation}

where $w_{p}$,  $w_{m}$, and  $w_{d}$ mean the weights assigned to the emotional scores from the poster, main soundtrack, and movie description, respectively. We selected the following weights based on subjective observations: $w_{p}$ = 1, $w_{m}$ = 2, $w_{d}$ = 3.

\subsection{Recommendation using Jaccard Coefficient}
\begin{figure}[]
    \centering
    \includegraphics[width=\linewidth]{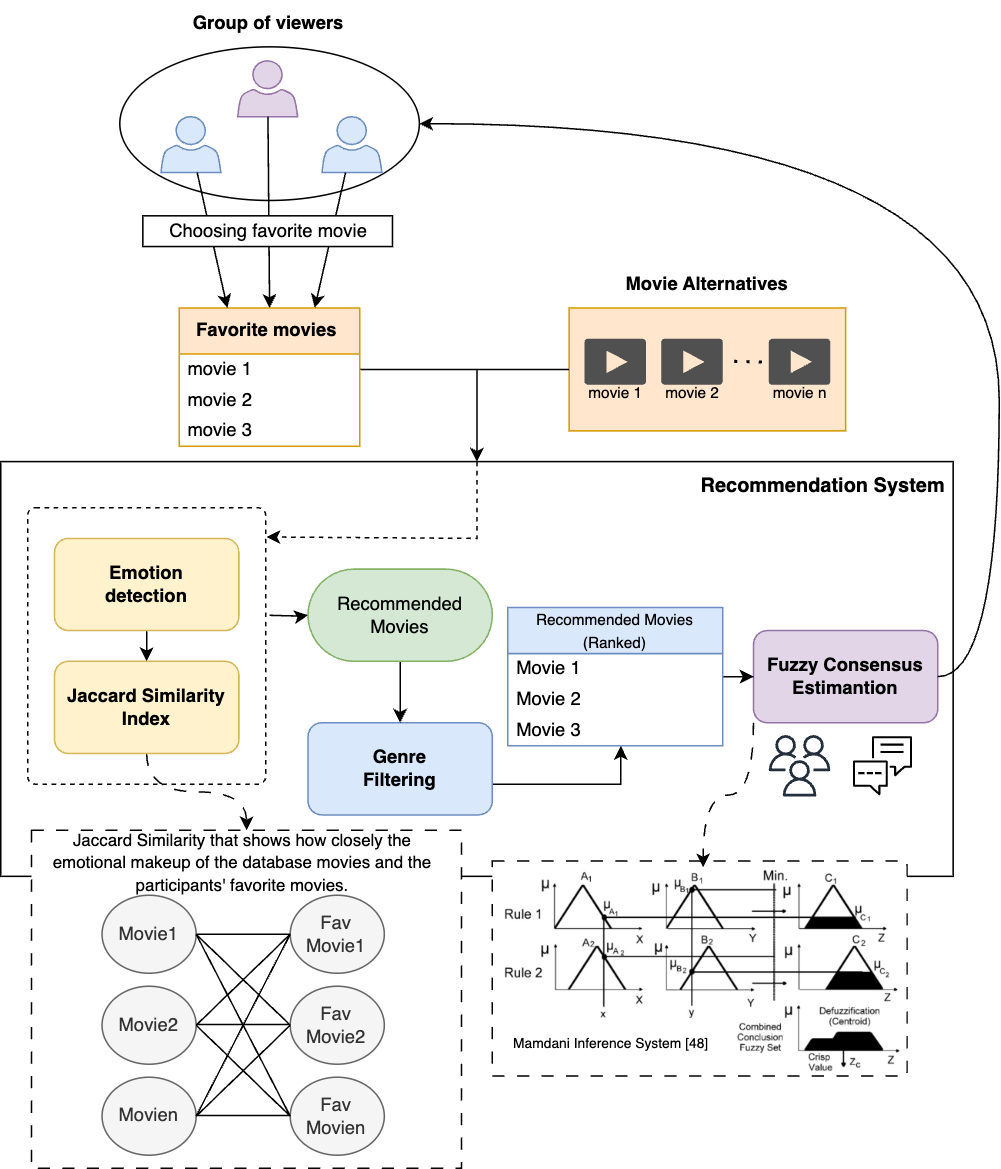}
    \caption{Recommendations are provided based on detected emotions and the Jaccard Similarity index, and then they are ordered by genre. Finally fuzzy inference system is used to evaluate consensus among group members regarding regarding recommended movies}
    \label{recsystem}
\end{figure}

We computed the Jaccard similarity coefficient between the emotional composition of movies in our database, and the participants' most loved movies using equation \eqref{jaccard}. This process gives a Jaccard value indicating the similarity between the emotional composition of the $n$ movies and the participants' preferred choices. Additionally, we can evaluate the effectiveness of our method by comparing the similarity between the actual and predicted ratings.

\begin{equation} \label{jaccard}
J(A, B) = \frac{\left|A\cap B \right|}{\left|A\cup B \right|}
\end{equation}

where  $J$ = Jaccard distance, $A$ = set 1, $B$ = set 2. Emotion was added with a threshold of 0.05 for the emotion score.


For group recommendations, we aggregate the Jaccard values obtained from each participant's favorite movie for each of the $n$ movies. The Jaccard value $J_{agg}^{i}$ for each movie is calculated as the average of Jaccard values across all participants. For movie $i$, the aggregated Jaccard value $J_{agg}^{i}$ is calculated by taking the average of Jaccard values obtained from participant preferences, $J_{j}^{i}$. 

\begin{equation} \label{3}
J_{agg}^{i} =  \frac{\sum_{j=1}^{m}J_{j}^{i}}{m}
\end{equation}

where $m$ denotes the total number of participants, $j$ - from 1 to $m$, $i$ - from 1 to $n$.

The proposed recommendation system can be seen in Fig. \ref{recsystem}. As can be seen, the recommended alternatives are then ranked according to their genre.


\subsection{Consensus Model using Fuzzy Logic}



\begin{figure}[tb]
    \centering
    \includegraphics[width=0.5\textwidth]{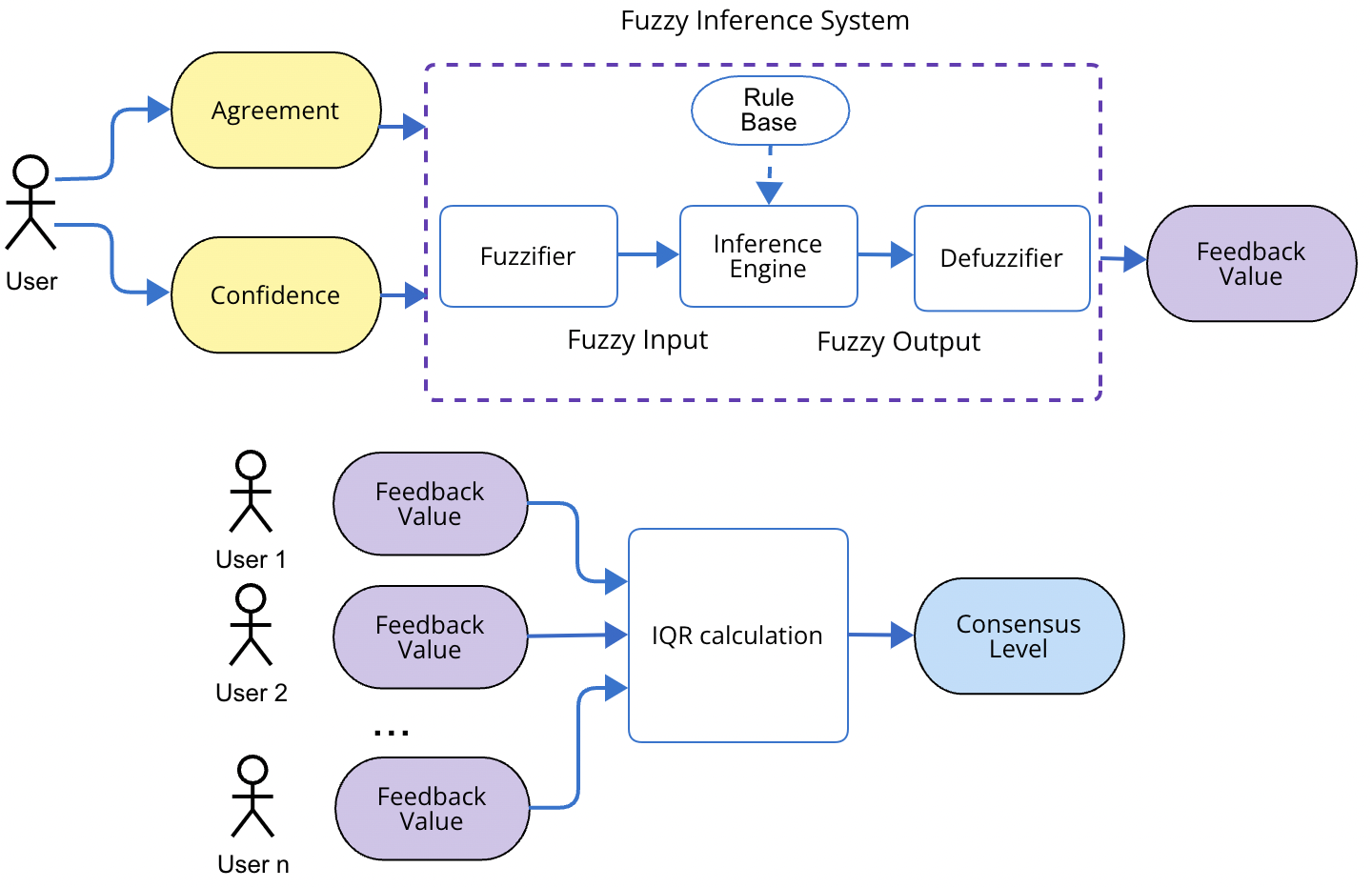}
    \caption{General process of consensus evaluation}
    \label{consensus}
\end{figure}

\begin{figure*}[tb]
    \centering
    \begin{subfigure}[t]{0.5\textwidth}
        \centering
        \includegraphics[height=1.7in]{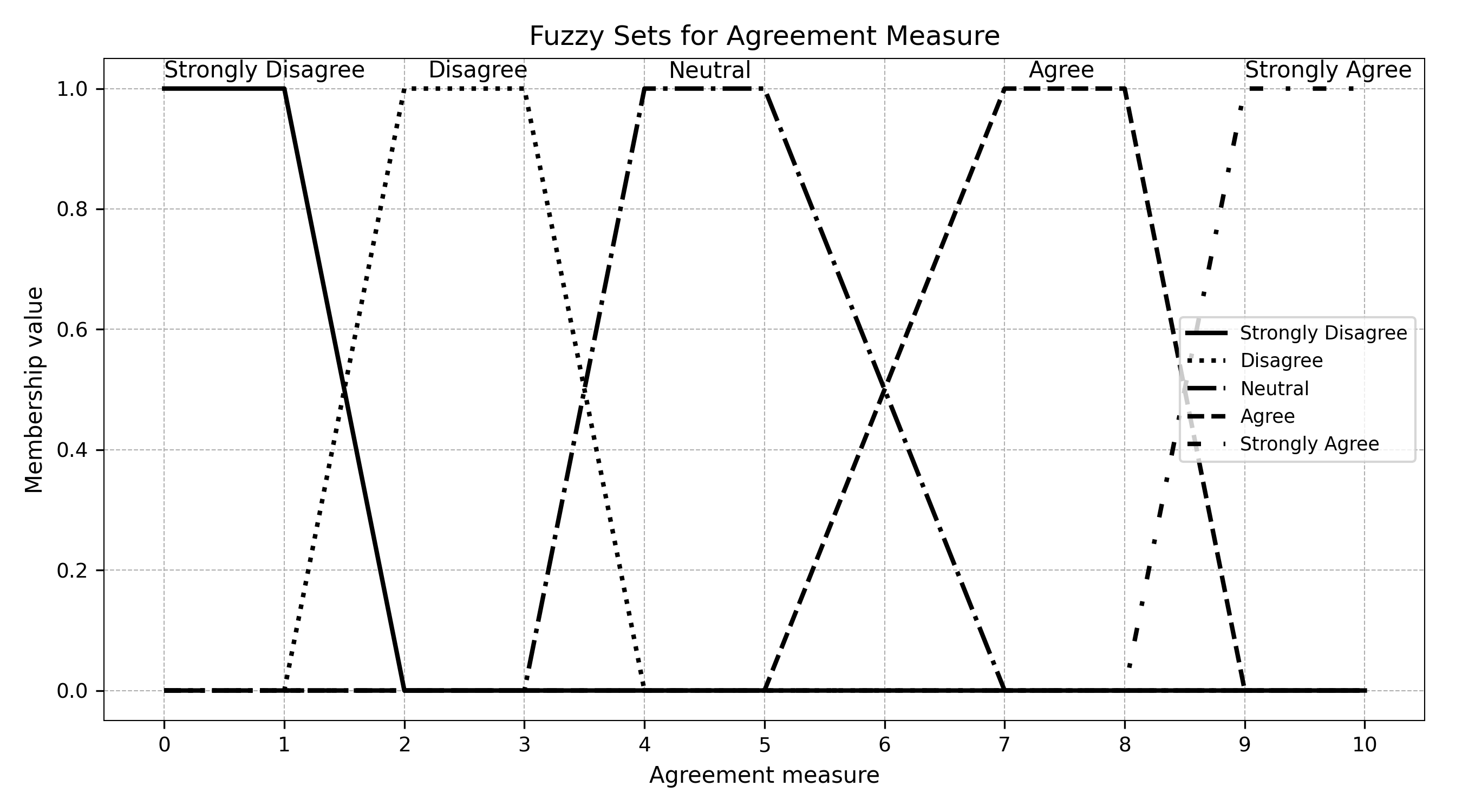}
        \caption{agreement measure}
        \label{input_mf:agr}
    \end{subfigure}%
    ~ 
    \begin{subfigure}[t]{0.5\textwidth}
        \centering
        \includegraphics[height=1.7in]{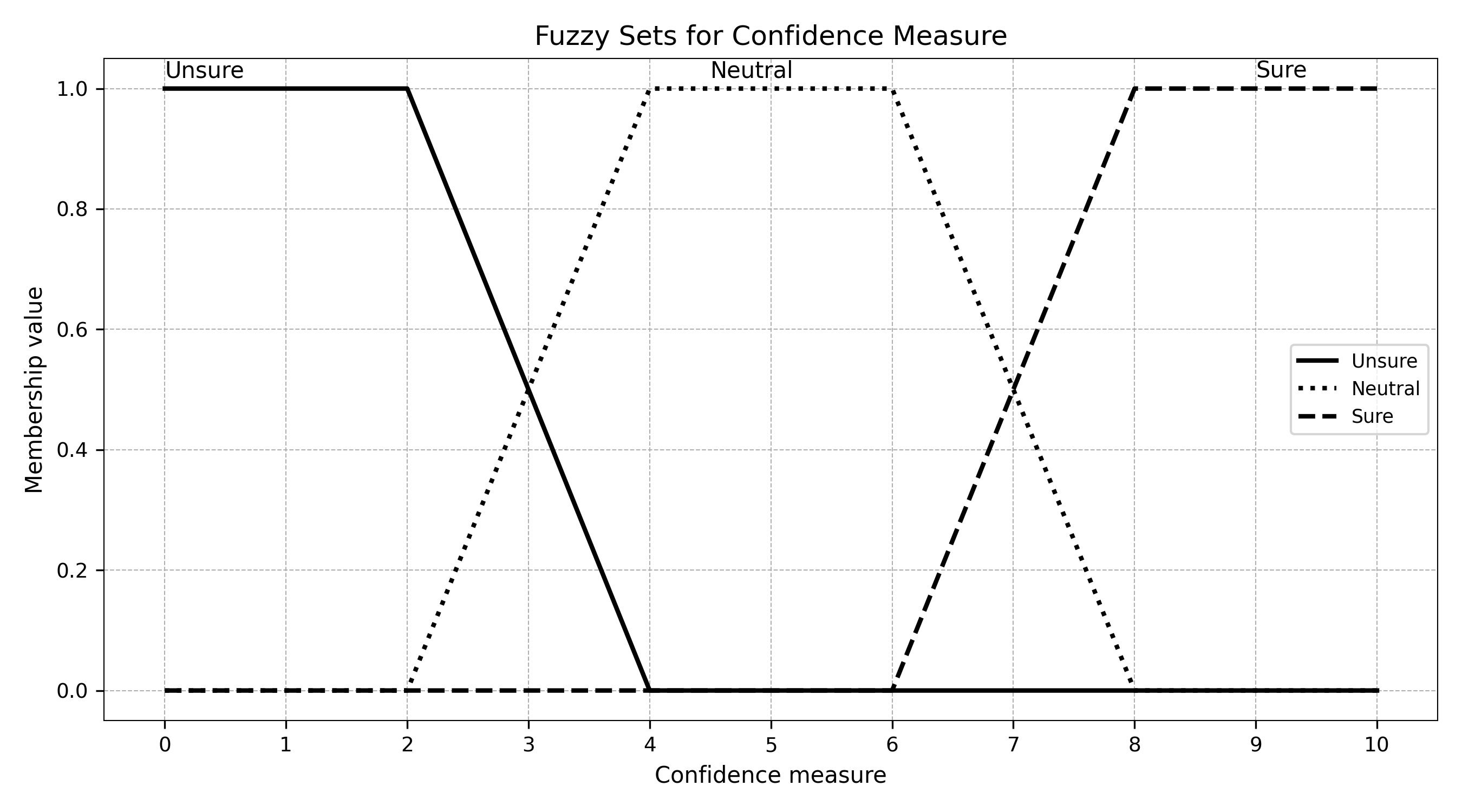}
        \caption{confidence measure}
        \label{input_mf:conf}
    \end{subfigure}
    \caption{Input Fuzzy Sets}
    \label{input_mf}
\end{figure*}

\begin{figure}[tb]
    \centering
    \includegraphics[width=0.4\textwidth, height=40mm]
    {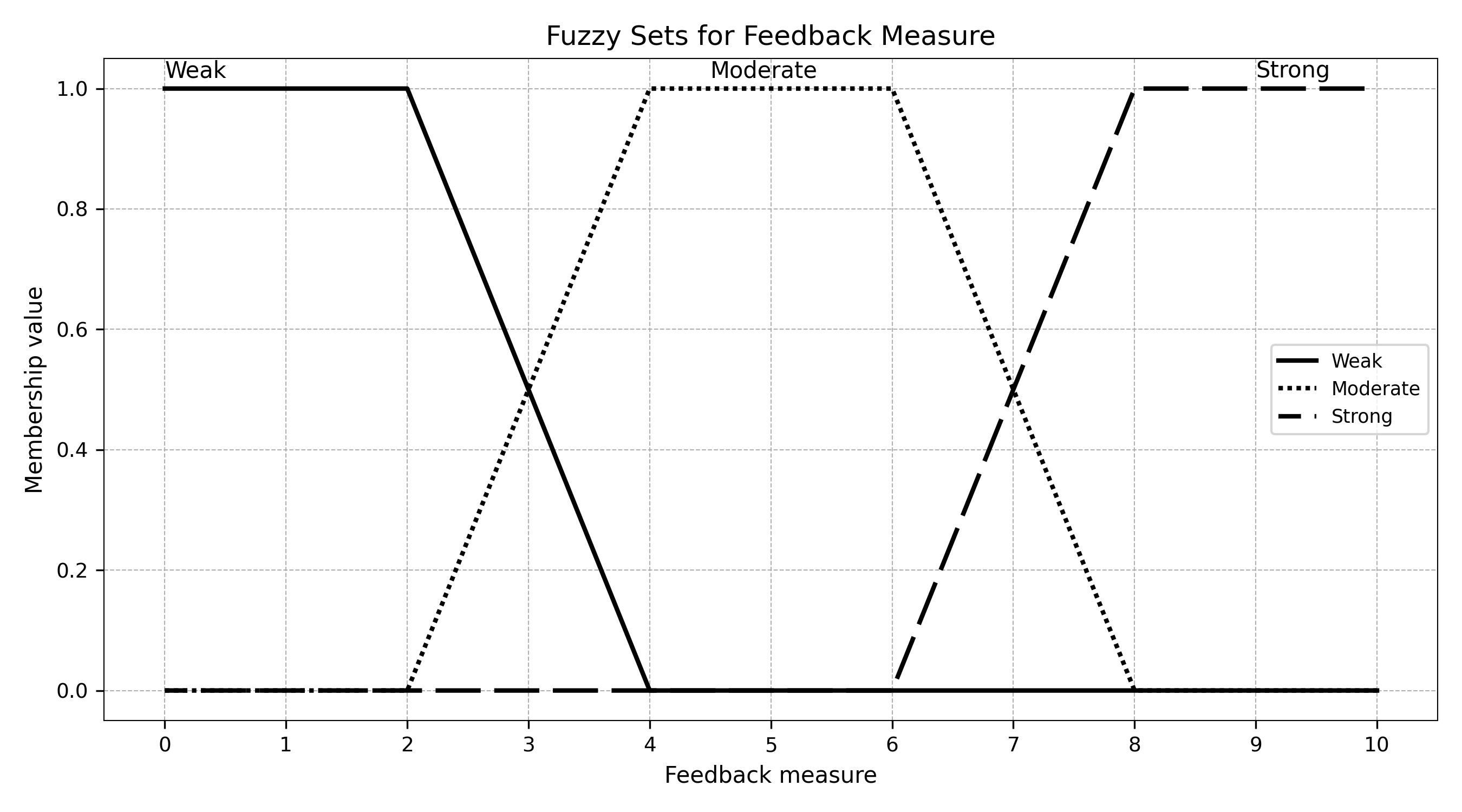}
    \caption{Output fuzzy set (feedback measure)}
     \label{output_mf}
\end{figure}

Group decision-making systems \cite{Enrique2002}, \cite{Herrera1996} often face challenges in achieving a high level of consensus among participants with diverse preferences \cite{fuzzy_consensus}, \cite{fuzcon}. In the context of movie recommendations, this task becomes particularly complex due to the subjective nature of movie appreciation. 
\\
After determining the most recommended movie for the group, the level of consensus \cite{mamdani} with the proposed movie is calculated considering participants' feedback. We propose a system that utilizes fuzzy logic to better accommodate and interpret the qualitative feedback of participants, thus aiming to enhance consensus and satisfaction with the recommended choice as illustrated in Fig \ref{consensus}.

Fuzzy logic  \cite{Zadeh1988} is a mathematical framework for dealing with uncertainty and imprecision, extending classical boolean logic by allowing truth values to range between 0 and 1. This approach facilitates the modeling of complex systems by enabling more nuanced decision-making and inference under conditions of ambiguity. By considering participants' feedback and confidence levels in these agreements, we aim to calculate the level of consensus measure as shown in Fig \ref{consensus}, using a Fuzzy inference system.

Our fuzzy logic inference system consists of two input fuzzy variables — \textit{Participant's Agreement} and \textit{Confidence Level}, and one fuzzy output variable - the \textit{Consensus Measure}:
\begin{enumerate}
    \item \textit{Participant's Agreement}: defined by linguistic terms such as "Highly Agree", "Agree", "Neutral", "Disagree", and "Strongly Disagree", converted into fuzzy values (see Fig. \ref{input_mf:agr}).
    \item \textit{Confidence Level}: defined by three fuzzy sets including "Unsure", "Neutral", and "Sure" (see Fig. \ref{input_mf:conf}). 
    \item \textit{Feedback Measure}: quantifies the group's overall consensus, with fuzzy sets representing "Weak", "Moderate", and "Strong" levels of feedback measure (Fig. \ref{output_mf}).
\end{enumerate}

The inference system uses a set of fuzzy rules to determine the consensus measure from the combination of agreement and confidence levels, as shown in Table \ref{tab:fuzzy_rules}.


This process ensures that the recommended movies closely align with the group's collective preferences and that participants are confident in these recommendations. Considering both the participants' agreement and confidence levels, our system offers a more sophisticated and accurate means of achieving consensus.

\begin{table}[tb]
\caption{Fuzzy rules for the fuzzy inference system}
\resizebox{\columnwidth}{!} {
    \begin{tabular}{|l|l|l|l|}
    \hline
    \textbf{Rule \#} & \textbf{Agreement} & \textbf{Confidence} & \textbf{Feedback value} \\ \hline
    1                & Strongly Agree     & Unsure              & Moderate                \\ \hline
    2                & Strongly Agree     & Neutral             & Strong                  \\ \hline
    3                & Strongly Agree     & Sure                & Strong                  \\ \hline
    4                & Agree              & Unsure              & Moderate                \\ \hline
    5                & Agree              & Neutral             & Moderate                \\ \hline
    6                & Agree              & Sure                & Strong                  \\ \hline
    7                & Neutral            & Unsure              & Moderate                \\ \hline
    8                & Neutral            & Neutral             & Moderate                \\ \hline
    9                & Neutral            & Sure                & Strong                  \\ \hline
    10               & Disagree           & Unsure              & Moderate                \\ \hline
    11               & Disagree           & Neutral             & Moderate                \\ \hline
    12               & Disagree           & Sure                & Weak                    \\ \hline
    13               & Strongly Disagree  & Unsure              & Moderate                \\ \hline
    14               & Strongly Disagree  & Neutral             & Weak                    \\ \hline
    15               & Strongly Disagree  & Sure                & Weak                    \\ \hline
    \end{tabular}
    \label{tab:fuzzy_rules}
    }
\end{table}

In our fuzzy logic system, we use trapezoidal membership functions for defining the fuzzy sets for variables "Participant's Agreement," "Confidence Level," described in Fig. \ref{input_mf} and "Feedback Measure" presented in Fig. \ref{output_mf}. Fuzzy set \cite{Zadeh1965} is defined as a set in which each element has a degree of membership that ranges between 0 and 1, indicating the degree to which the element belongs to the set.
The trapezoidal membership function is one of the standard shapes used in fuzzy logic systems due to its flexibility and simplicity. It is defined by four points, creating a shape that resembles a trapezoid. The parameters $a$ and $d$ control the trapezoid's left and right feet or base points, and the parameters $b$ and $c$ control the trapezoid's left and right shoulders or top points.
The general form of a trapezoidal membership function $\mu(x)$ can be described mathematically as:

\begin{equation}
\mu(x) = \max\left(\min\left(\frac{x - a}{b - a}, 1, \frac{d - x}{d - c}\right), 0\right)
\label{trapezoidal}
\end{equation}

The final consensus measure for the group was evaluated by using the measure of statistical dispersion - Interquartile Range (IQR) (\ref{iqr}) and the measure of central tendency - Mean (\ref{mean_val}) inference across all participants to obtain a crisp value indicating the group's overall consensus, in Table \ref{consensus_level_meth}.

\begin{equation}\label{iqr}
    IQR = Q_{3} - Q_{1}
\end{equation}
where $Q_{1}$ is the first quartile, $Q_{3}$ is the third quartile of distribution.  

\begin{equation}\label{mean_val}
\text{Mean} = \frac{\sum_{i=1}^{n} x_i}{n}
\end{equation}
where $n$ is the total number of participants, $i$ - from 1 to $n$, $x_i$ - Feedback Measure of $i$-th participant.

\begin{table}[tb]
\caption{Consensus level}
\centering
\begin{tabular}{|c|l|}
\hline
\textbf{\begin{tabular}[c]{@{}c@{}}Interquartile Range \\ (IQR)\end{tabular}} & \multicolumn{1}{c|}{\textbf{Consensus level}} \\ \hline
0.00 - 2.00                                                                   & High                                          \\ \hline
2.01 - 4.00                                                                   & Medium                                        \\ \hline
\textgreater{}= 4.01                                                          & None                                          \\ \hline
\end{tabular}
    \label{consensus_level_meth}
\end{table}

A satisfactory consensus measure is defined by a level of IQR and mean score. When the calculated IQR exceeds the set threshold, or the calculated mean does not reach the corresponding set threshold value, it signals the need to re-evaluate the movie options or the recommendation process. Conversely, otherwise indicates a successful recommendation, which confirms that most of the group is satisfied with the movie's recommendation.




\section{Experimental Results}


\subsection{Performance Evaluation}

\begin{figure}[h]
    \centering
    \includegraphics[width=0.3\textwidth]{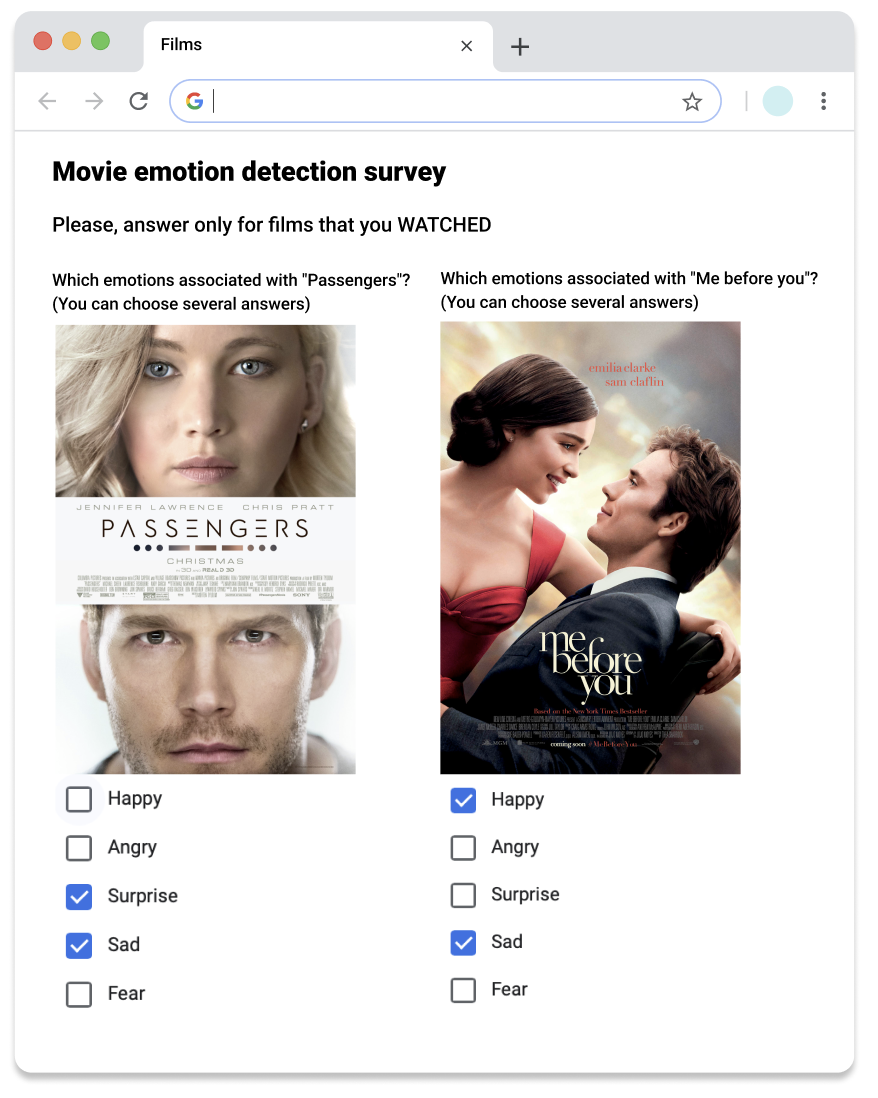}
    \caption{Survey on Movie Emotion Recognition}
    \label{survey_snap}
\end{figure}

We surveyed individuals to examine their emotional reactions to 12 movies shown in Table \ref{input_12_movies}, with responses categorized as \textit{Happy, Anger, Surprise, Sad, and Fear}. Respondents were allowed to provide more than one emotional reaction for each question and were asked to list only the movies they had seen and their relevant feelings while watching them (see Fig. \ref{survey_snap}).

\begin{table}[tb]
\caption{Movie alternatives for the experiments}
\resizebox{\linewidth}{!}{%
\begin{tabular}{|l|l|l|l|p{2cm}|}
\hline
id & Movie & Description text & Soundtrack & Poster \\ \hline
1 & Insidious 3 & \begin{tabular}[c]{@{}l@{}}
    After trying to connect \\ with her dead mother, \\ teenager Quinn Brenner, \\ asks  psychic Elise ...\end{tabular} & \begin{tabular}[c]{@{}l@{}}The Insidious \\ Plane by \\Joseph \\Bishara\end{tabular} & \begin{minipage}{.2\textwidth} \includegraphics[width=0.4\linewidth, height=18mm]{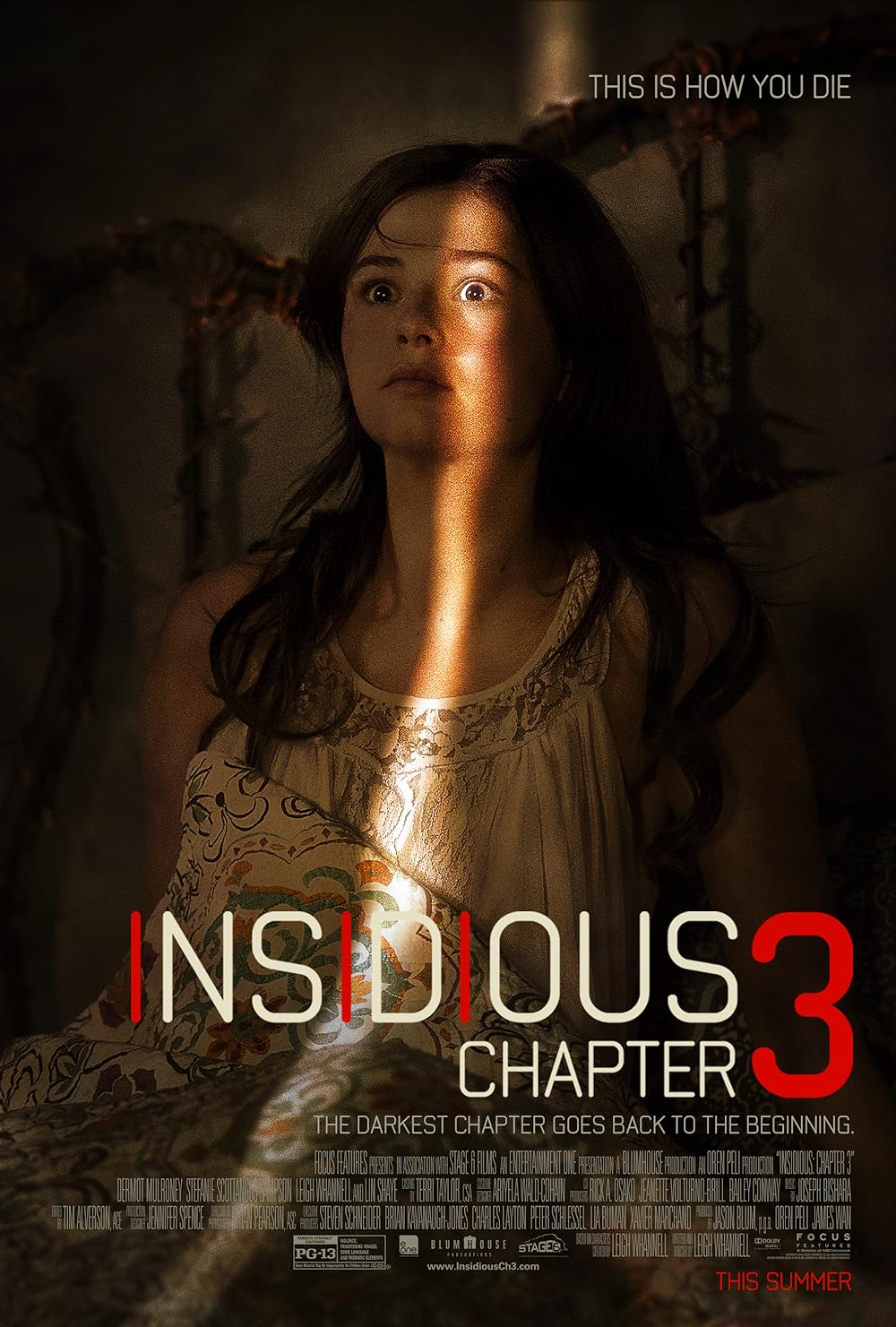}
    \end{minipage}\\ \hline
2 & \begin{tabular}[c]{@{}l@{}}Annabelle:\\ Creation\end{tabular} & \begin{tabular}[c]{@{}l@{}}
          Doll manufacturer Samuel \\ Mullins is a happy \\ family man with his \\ wife Esther and \\ their daughter... \end{tabular} & \begin{tabular}[c]{@{}l@{}}Somethin’\\ Special by\\ Colbie\\ Caillat\end{tabular} & \begin{minipage}{.2\textwidth} \includegraphics[width=0.4\linewidth, height=18mm]{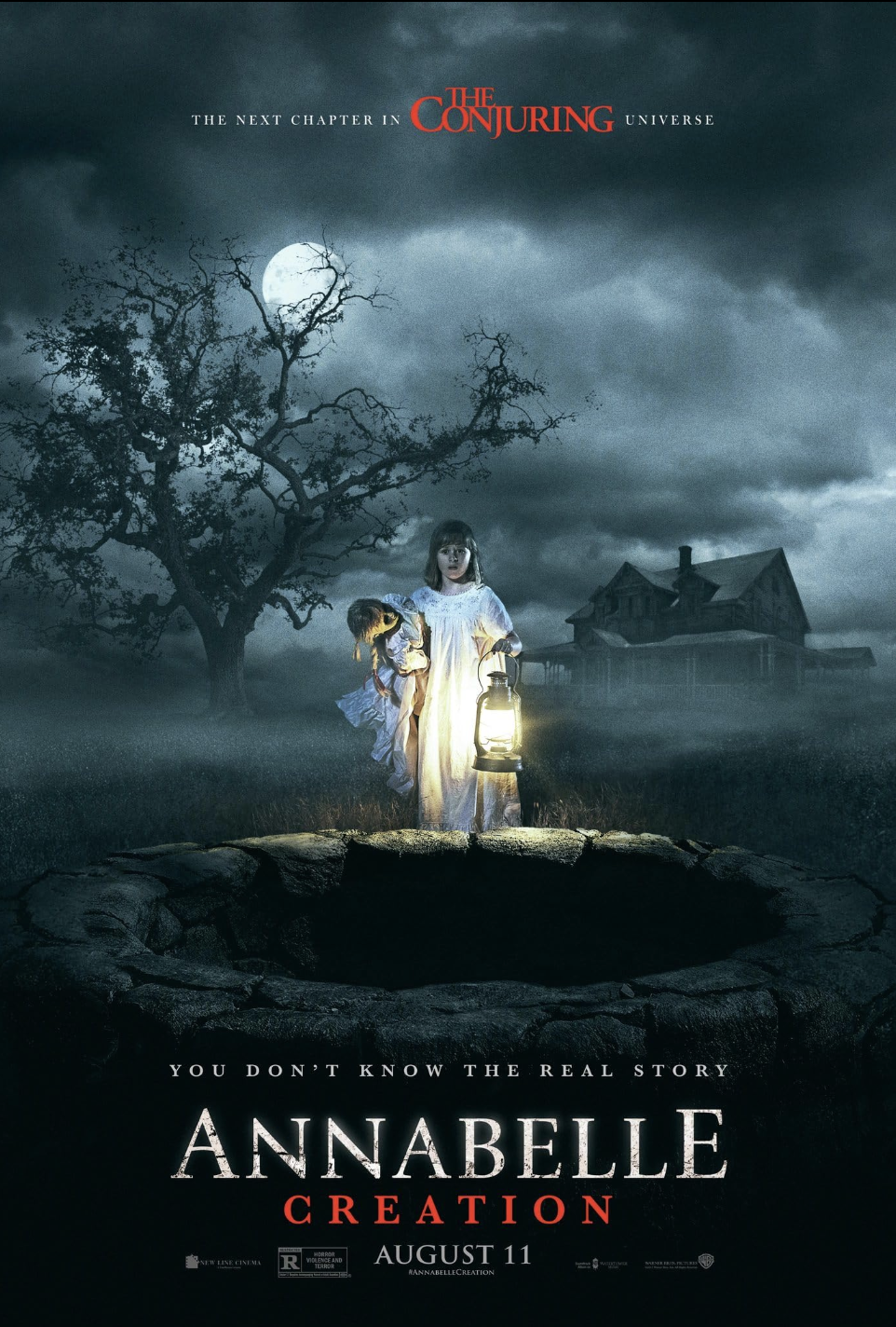}
    \end{minipage}\\ \hline
… & … & … & … & … \\ \hline
12 & \begin{tabular}[c]{@{}l@{}}Me before\\ you\end{tabular} & \begin{tabular}[c]{@{}l@{}} Lou Clark knows \\ lots of things. She \\ knows how many \\ footsteps there are \\ between the ... \end{tabular} & \begin{tabular}[c]{@{}l@{}}Numb \\ by Max\\ Jury\end{tabular} &  \begin{minipage}{.2\textwidth} \includegraphics[width=0.4\linewidth, height=18mm]{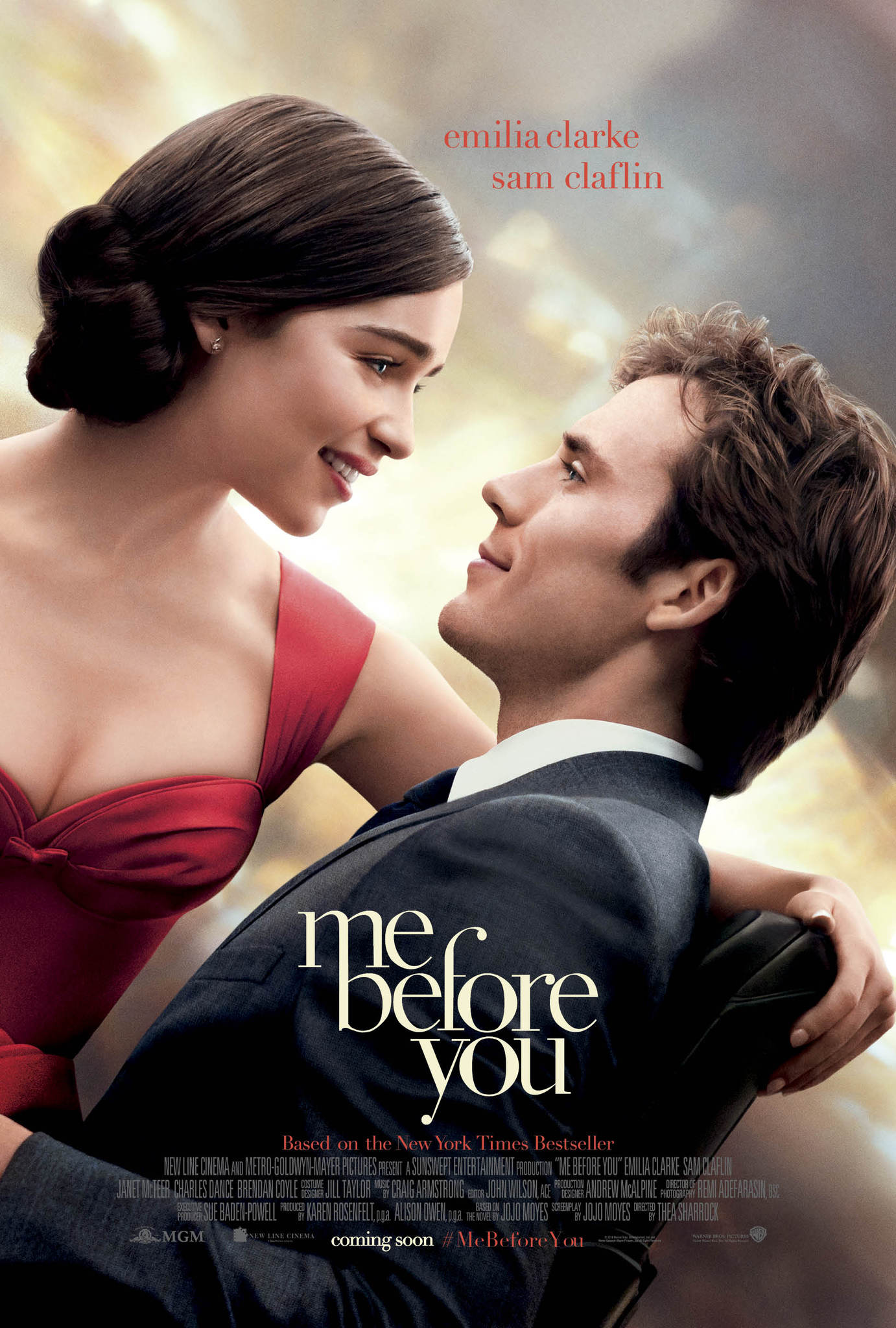}
    \end{minipage}\\ \hline
\end{tabular}
}
\label{input_12_movies}
\end{table}

\begin{figure}[h]
    \centering
    \includegraphics[width=0.5\textwidth]{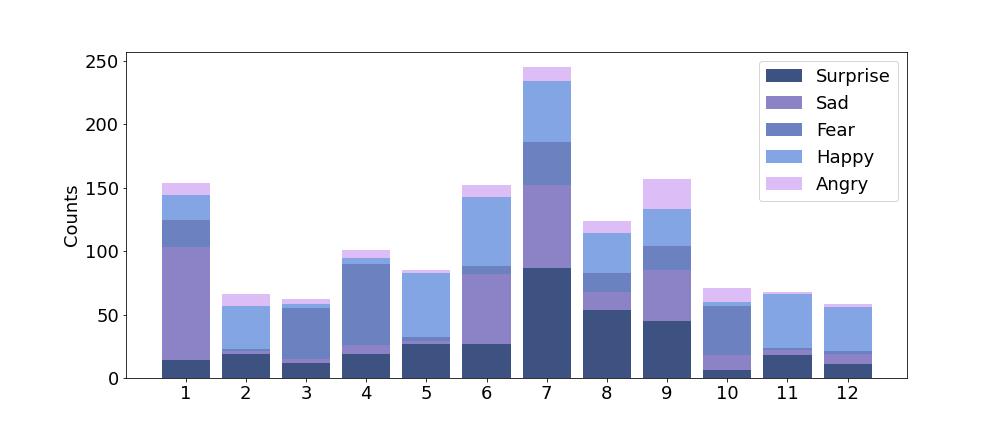}
    \caption{Survey results' distribution}
    \label{survey_results}
\end{figure}

As a result, we obtained human ratings for emotion distribution for each of the 12 movies, illustrated in Fig. \ref{survey_results}. This bar chart shows the degree of agreement amongst participants. For instance, looking at the responses for the movie "Just go with it" (number 5), we can see that most people selected \textit{Happy} and \textit{Surprise}, nobody selected \textit{Fear}, and some selected \textit{Sad} and \textit{Angry}. It can be an example of the best consensus of the group. In contrast, answers for the movie "Interstellar" (number 7) range widely, involving all five emotions, and four of them are chosen by a significant portion of respondents, making it the worst consensus example.

\begin{figure}[h]
    \centering
    \begin{subfigure}[b]{0.5\textwidth}
        \centering
        \includegraphics[height=1.3in]{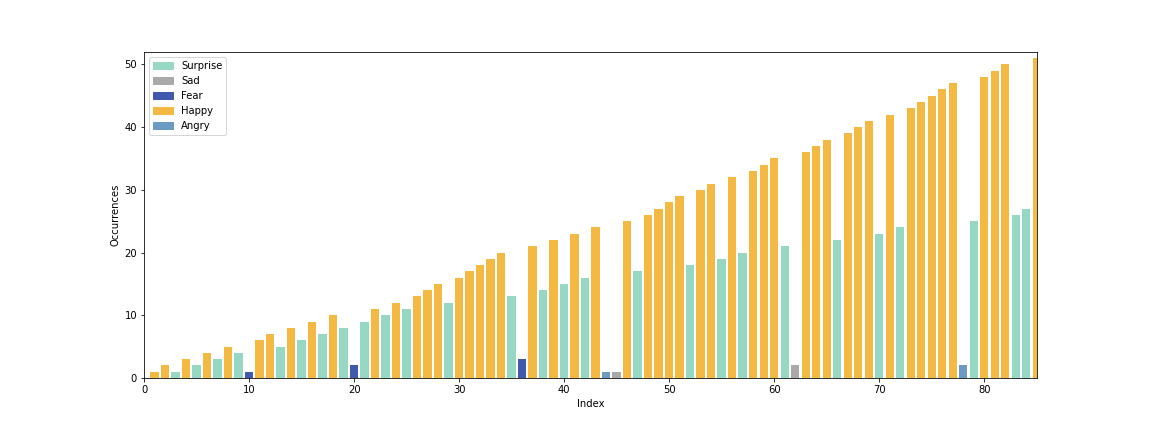}
        \caption{The best consent between participants (the movie "Just go with it")}
        \label{best}
    \end{subfigure}%
    \vfill
    \begin{subfigure}[b]{0.5\textwidth}
        \centering
        \includegraphics[height=1.3in]{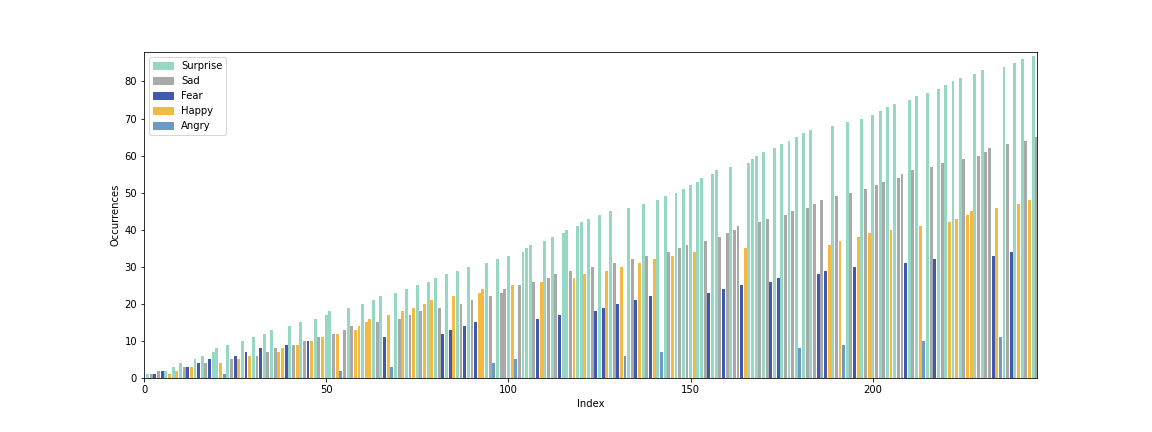}
        \caption{The worst consent between participants for the movie "Interstellar"}
        \label{worst}
    \end{subfigure}
    \caption{Consensus between survey participants regarding the prevailing emotions in movies}
    \label{consensus_survey}
\end{figure}

Fig. \ref{consensus_survey} represents survey results for two movies achieving the participants' best and worst consent levels. The fifth movie's replies are displayed in a histogram under \ref{best}, where the dominating emotions and obvious outliers are evident. The seventh movie's answers are included under \ref{worst}, highlighting the severe discrepancies in response patterns.

The specific emotion score for a particular movie was calculated as the proportion of selection of this emotion among all choices made for this movie. The results are shown in Table \ref{survey_emotion}. Using data from three separate channels, the predicted scores for \textit{Happiness, Anger, Surprise, Sadness,} and \textit{Fear} are combined using \eqref{agg}. The results are shown in Table \ref{predicted_scores}.

\begin{table}[tb]
\caption{Emotion scores of survey participants}
\resizebox{\linewidth}{!}{%
\begin{tabular}{|l|l|l|}
\hline
id & Movie       & Emotion scores of survey participants                                                                                    \\ \hline
1  & Insidious 3     & \begin{tabular}[c]{@{}l@{}}\{'Happy': 0.05, 'Angry': 0.09, 'Surprise': 0.23,\\  'Sad': 0.05, 'Fear': 0.59\}\end{tabular} \\ \hline
2  & Annabele: Creation  & \begin{tabular}[c]{@{}l@{}}\{'Happy': 0.0, 'Angry': 0.03, 'Surprise': 0.26,\\ 'Sad': 0.06, 'Fear': 0.65\}\end{tabular} \\ \hline
…  & …           & …                                                                                                                       \\ \hline
12 & Me before you & \begin{tabular}[c]{@{}l@{}}\{'Happy': 0.35, 'Angry': 0.09, 'Surprise': 0.17,\\ 'Sad': 0.3, 'Fear': 0.09\}\end{tabular} \\ \hline
\end{tabular}}
\label{survey_emotion}
\end{table}

\begin{figure*}[h]
    \centering
    \begin{subfigure}[t]{0.5\textwidth}
        \centering
        \includegraphics[height=1.8in]{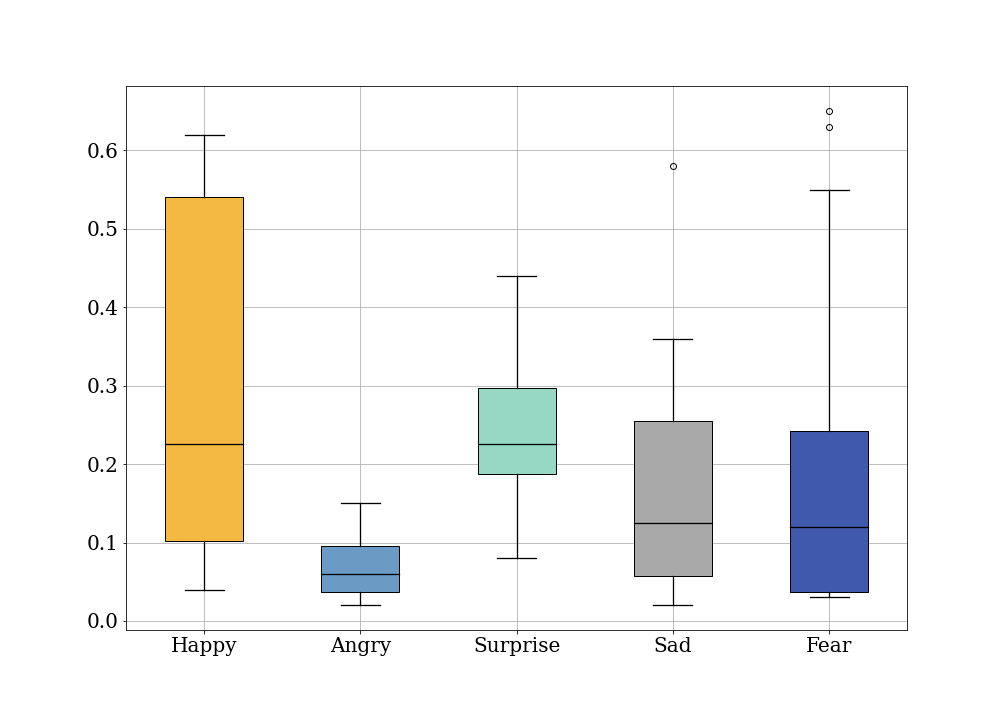}
        \caption{Human Ratings}
        \label{boxplot:a}
    \end{subfigure}%
    \hfill
    \begin{subfigure}[t]{0.5\textwidth}
        \centering
        \includegraphics[height=1.8in]{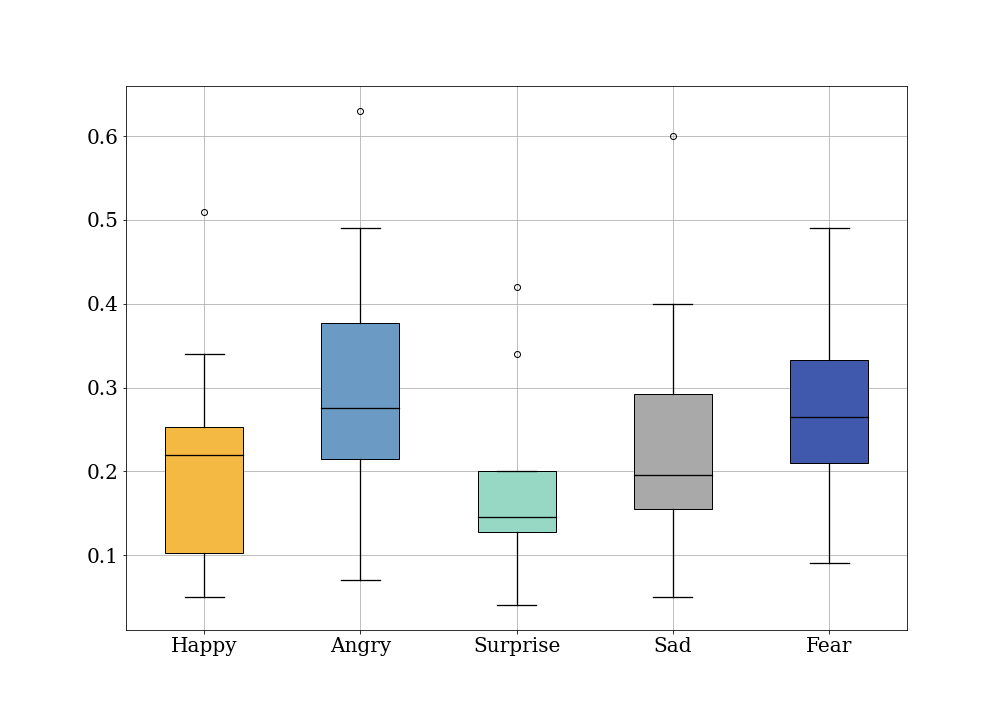}
        \caption{Predicted Ratings}
        \label{boxplot:b}
    \end{subfigure}
    \caption{Emotion score distribution: human versus predicted ratings }
    \label{boxplot}
\end{figure*}

\begin{table}[t]
\caption{Input parameters of participants’ favorite movies
}
\label{fav_movies}
\centering
\begin{adjustbox}{center}
\resizebox{\linewidth}{!}{%
\begin{tabular}{|l|l|l|l|p{2cm}|}
\hline
id &
  Movie &
  Description text &
  \begin{tabular}[c]{@{}l@{}}Soundtrack\end{tabular} &
  Poster \\ \hline
1 &
  \begin{tabular}[c]{@{}l@{}}The\\ Notebook\end{tabular} &
  \begin{tabular}[c]{@{}l@{}}With almost religious\\ devotion, Duke, a\\ kind octogenarian…\end{tabular} &
  \begin{tabular}[c]{@{}l@{}}I'll Be\\ Seeing You\\ by Billie\\ Holiday\end{tabular} & \begin{minipage}{.2\textwidth} \includegraphics[width=0.4\linewidth, height=17mm]{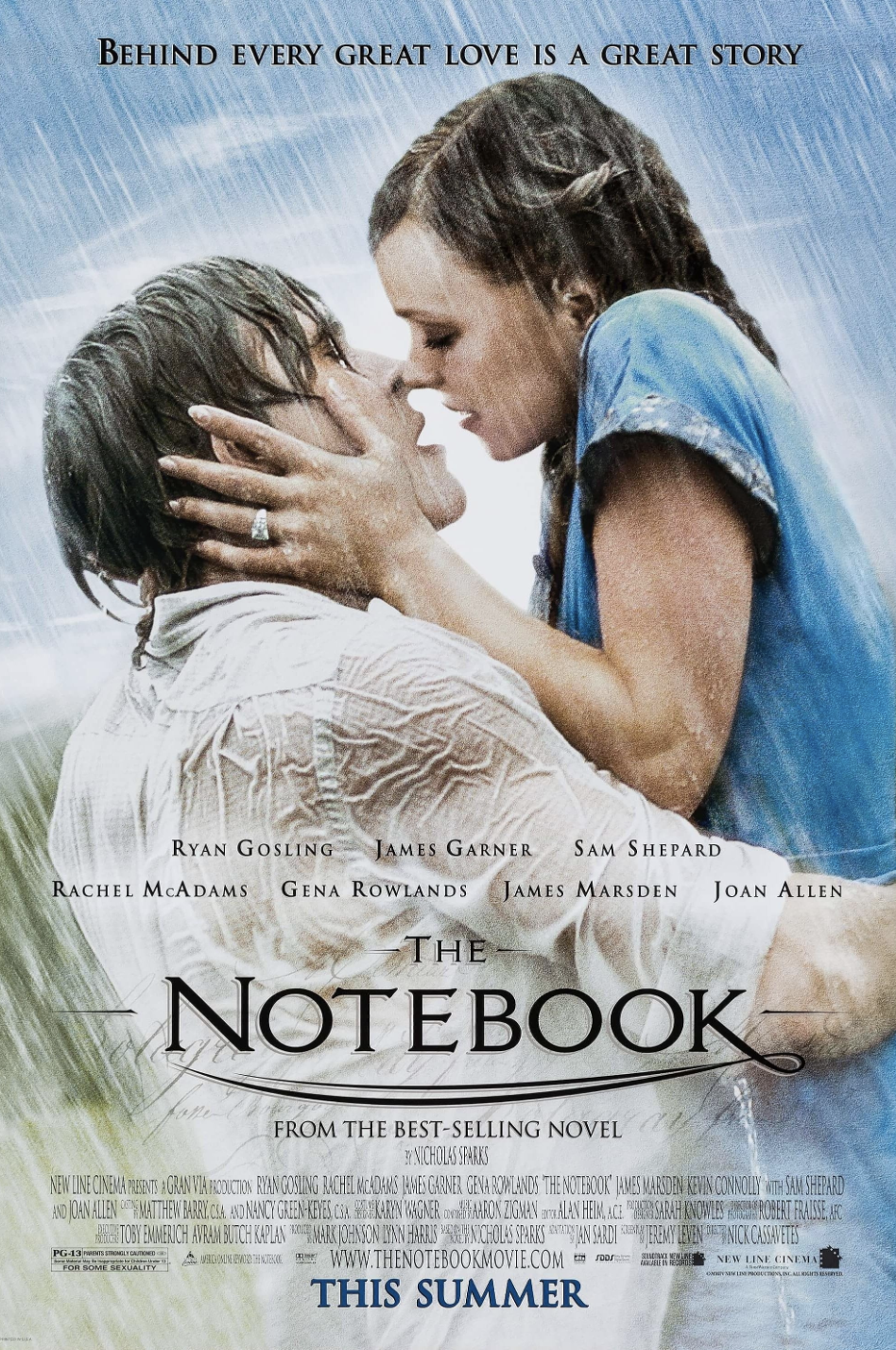}
    \end{minipage}
   \\ \hline
2 &
  Split &
  \begin{tabular}[c]{@{}l@{}}Though Kevin\\ (James McAvoy) has\\ evidenced 23\\ personalities to his…\end{tabular} &
  \begin{tabular}[c]{@{}l@{}}In\\ September\\ by Slam\\ Allen\end{tabular} & \begin{minipage}{.2\textwidth} \includegraphics[width=0.4\linewidth, height=17mm]{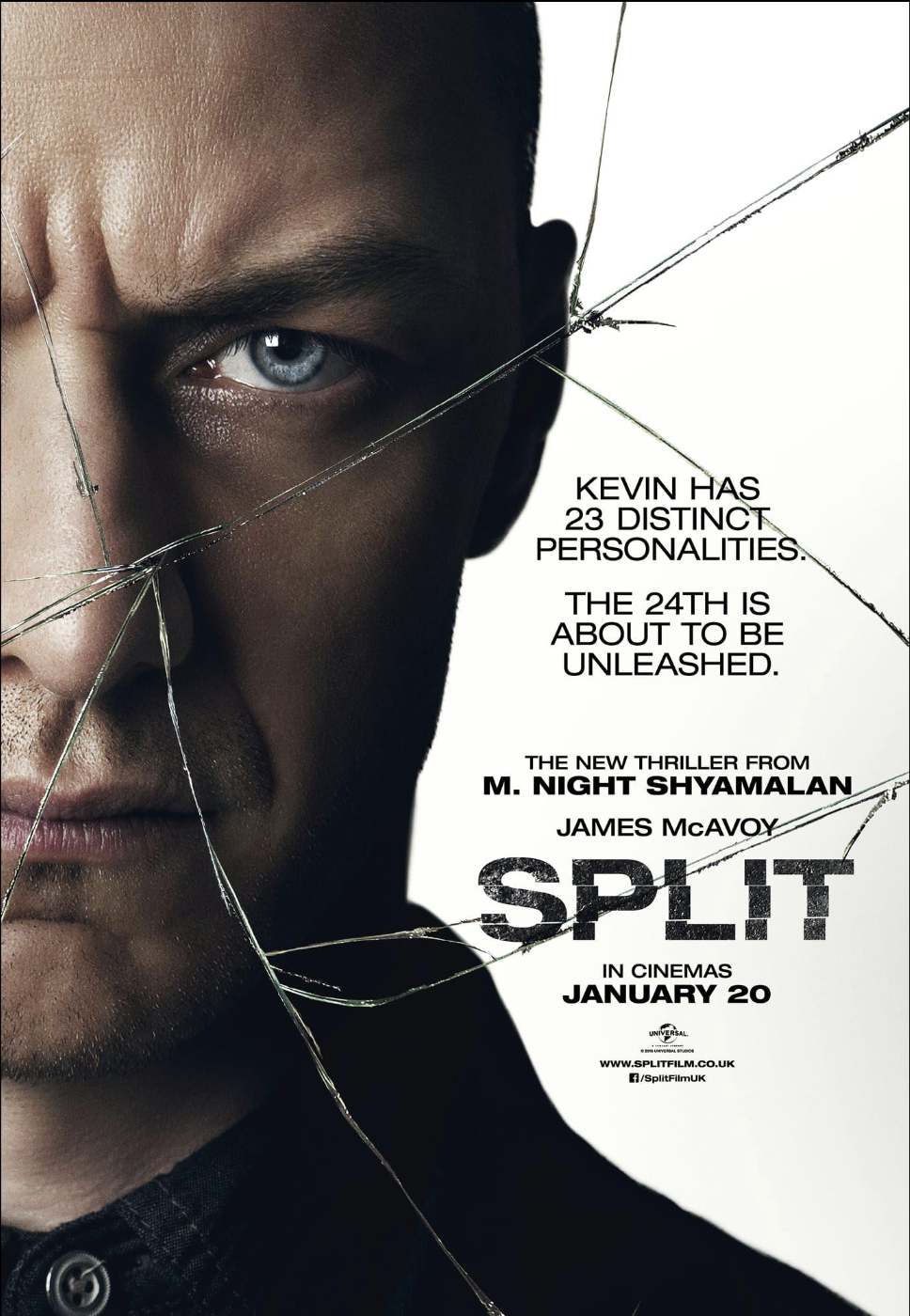}
    \end{minipage}
   \\ \hline
3 &
  Oppenheimer &
  \begin{tabular}[c]{@{}l@{}}A dramatization of\\ the life story of J.\\ Robert Oppenheimer,\\ the physicist who had…\end{tabular} &
  \begin{tabular}[c]{@{}l@{}}Can You\\ Hear The\\ Music by\\ Ludwig\\ Göransson\end{tabular} & \begin{minipage}{.2\textwidth} \includegraphics[width=0.4\linewidth, height=17mm]{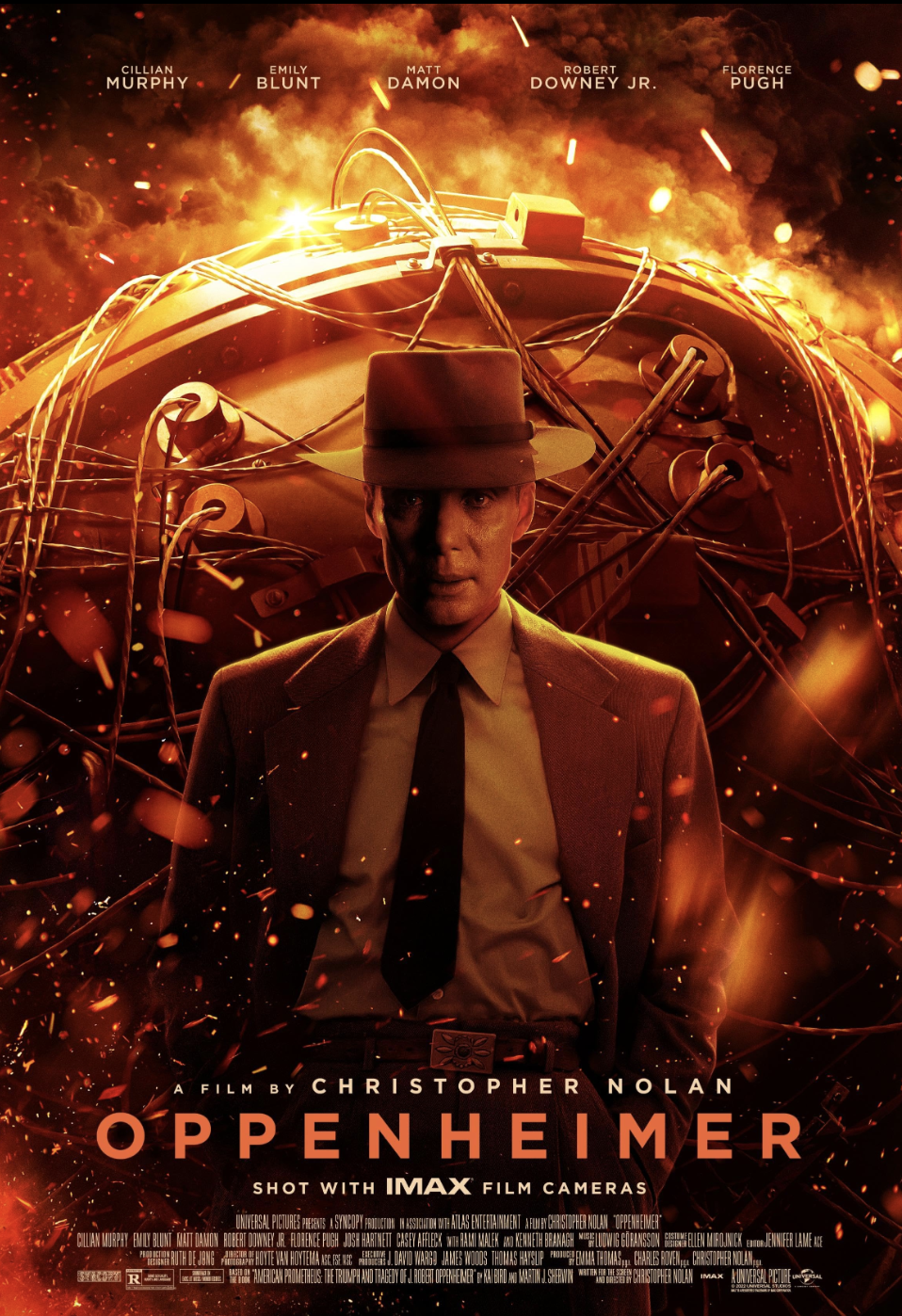}
    \end{minipage}
   \\ \hline
4 &
  Barbie &
  \begin{tabular}[c]{@{}l@{}}Barbie the Doll lives\\ in bliss in the\\ matriarchal society of\\ Barbieland feeling…\end{tabular} &
  \begin{tabular}[c]{@{}l@{}}Dance the\\ Night by\\ Dua Lipa\end{tabular} & \begin{minipage}{.2\textwidth} \includegraphics[width=0.4\linewidth, height=17mm]{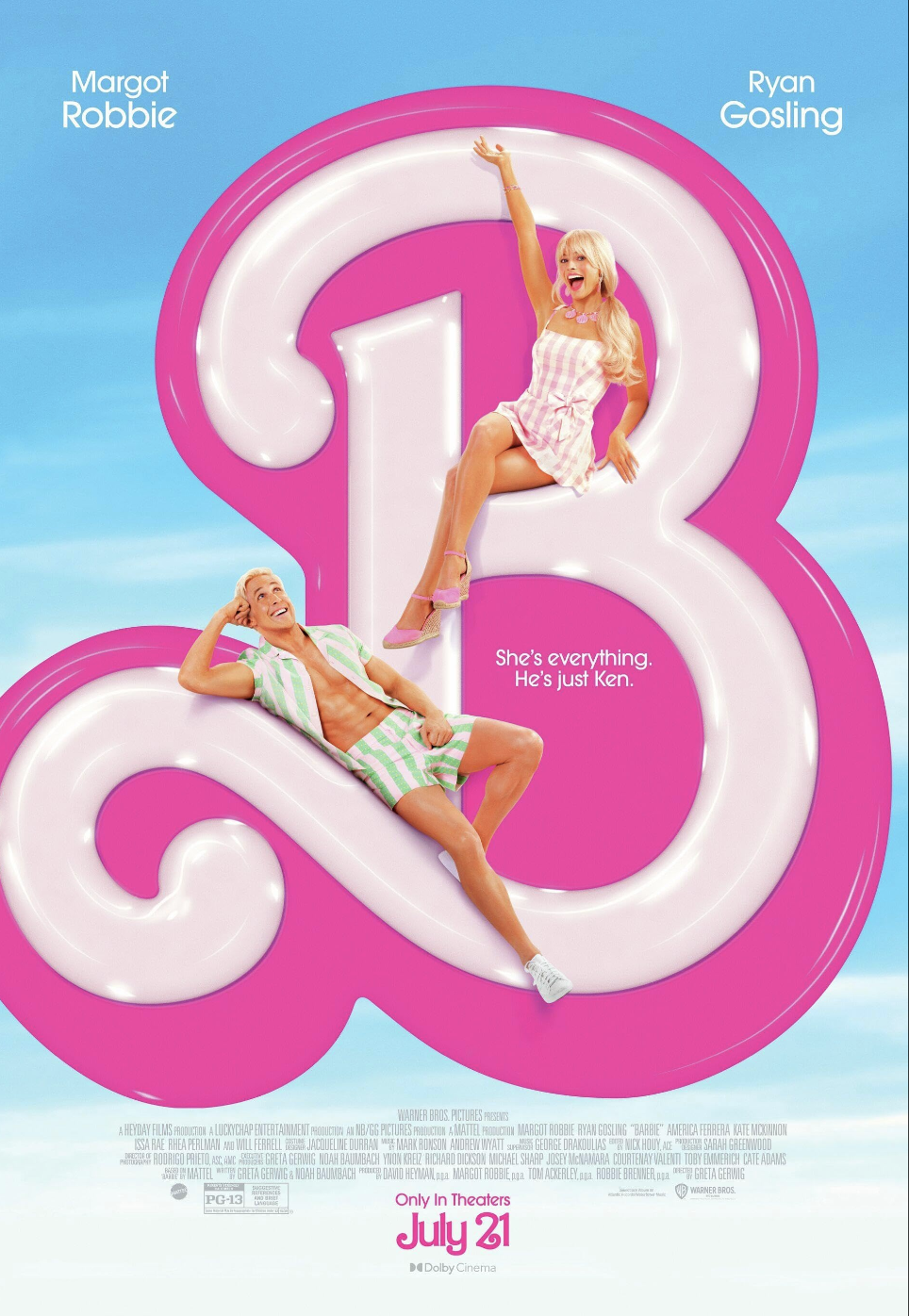}
    \end{minipage}
   \\ \hline
\end{tabular}}
\end{adjustbox}
\label{input_4_movies}
\end{table}

Fig. \ref{boxplot} illustrates real (Fig. \ref{boxplot:a}) and predicted   (Fig. \ref{boxplot:a}) emotion score distributions. Analyzing the distribution of the emotion scores for our predicted values (Fig. \ref{boxplot:a}), we observed that each emotion has a minimum significance of the first quartile(Q1) of almost 0.05. So,  the threshold was selected as 0.05.

We calculated the Pearson correlation coefficient between each emotion channel (text, colors, audio) and real human ratings to evaluate which emotion channel has the most significant impact on human impression. With a correlation of 0.43, the text description emotion channel and human ratings had the highest relationship.

Now, we can compare the prediction power of our approach by finding the similarity index between predicted emotion scores and the scores given by survey participants. Using the aggregated results of predicted scores (see Table \ref{predicted_scores}) and real human scores (Table \ref{survey_emotion}), we calculated the Jaccard distance \eqref{jaccard} between the real and the predicted values of emotions for each movie  (see Table \ref{jaccard_sim}). Afterward, we estimated the average similarity coefficient to be 0.76. 

\begin{table*}[t]
\centering
\caption{Jaccard Similarity Index between predicted and real emotion distribution for each movie. The mean Jaccard Similarity Index is 0.76}
\begin{adjustbox}{center}
\resizebox{\linewidth}{!}{%
\begin{tabular}{|l|l|l|l|l|l|l|l|l|l|l|l|l|}
\hline
Movie &
  Titanic &
  \begin{tabular}[c]{@{}l@{}}Bride\\ wars\end{tabular} &
  Insidious 3 &
  \begin{tabular}[c]{@{}l@{}}Annabelle:\\ Creation\end{tabular} &
  \begin{tabular}[c]{@{}l@{}}Just go\\ with it\end{tabular} &
  \begin{tabular}[c]{@{}l@{}}Me before\\ you\end{tabular} &
  Interstellar &
  \begin{tabular}[c]{@{}l@{}}Edge of\\ tomorrow\end{tabular} &
  Passengers &
  \begin{tabular}[c]{@{}l@{}}Don’t\\ breathe 2\end{tabular} &
  \begin{tabular}[c]{@{}l@{}}The\\ Proposal\end{tabular} &
  \begin{tabular}[c]{@{}l@{}}The\\ holiday\end{tabular} \\ \hline
\begin{tabular}[c]{@{}l@{}}Jaccard\\ Similarity\end{tabular} &
  1 &
  0.6 &
  0.4 &
  0.8 &
  0.5 &
  0.8 &
  0.8 &
  1.0 &
  1.0 &
  1.0 &
  0.6 &
  0.6 \\ \hline
\end{tabular}}
\end{adjustbox}
\label{jaccard_sim}
\end{table*}

\begin{table*}[t]
\centering
\caption{Predicted Emotion score of movies}
\begin{adjustbox}{center}
\resizebox{\linewidth}{!}{%
\begin{tabular}{|l|l|l|l|l|l|}
\hline
id &
  Movie &
  \begin{tabular}[c]{@{}l@{}}Emotion score of movies\\ from description text\end{tabular} &
  \begin{tabular}[c]{@{}l@{}}Emotion score of movies\\ from soundtrack\end{tabular} &
  \begin{tabular}[c]{@{}l@{}}Emotion score of movies\\ from poster\end{tabular} &
  \begin{tabular}[c]{@{}l@{}}Average emotion score\\ of movies\end{tabular} \\ \hline
1 &

  Insidious 3 &
  \begin{tabular}[c]{@{}l@{}}\{'Happy': 0, 'Angry':\\ 0.12, 'Surprise': 0.0, 'Sad':\\ 0.38, 'Fear': 0.5\}\end{tabular} &
  \begin{tabular}[c]{@{}l@{}}\{'Happy': 0.0, 'Angry':\\ 0.5, 'Surprise': 0.0, 'Sad':\\ 0.0, 'Fear': 0.5\}\end{tabular} &
  \begin{tabular}[c]{@{}l@{}}\{'Happy': 0.33, 'Angry':\\ 0.43, 'Surprise': 0.25,\\ 'Sad': 0.38, 'Fear': 0.33\}\end{tabular} &
  \begin{tabular}[c]{@{}l@{}}\{'Happy': 0.06, 'Angry': 0.3,\\ 'Surprise': 0.04, 'Sad': 0.25,\\ 'Fear': 0.47\}\end{tabular} \\ \hline
2 &
  Annabele: Creation &
  \begin{tabular}[c]{@{}l@{}}\{'Happy': 0.16, 'Angry':\\ 0.05, 'Surprise': 0.21, 'Sad':\\ 0.16, 'Fear': 0.42\}\end{tabular} &
  \begin{tabular}[c]{@{}l@{}}\{'Happy': 0, 'Angry':\\ 0.86, 'Surprise': 0, 'Sad':\\ 0, 'Fear': 0.14\}\end{tabular} &
  \begin{tabular}[c]{@{}l@{}}\{'Happy': 0.5, 'Angry':\\ 0.3, 'Surprise': 0.3,\\ 'Sad': 0.56, 'Fear': 0.5\}\end{tabular} &
  \begin{tabular}[c]{@{}l@{}}\{'Happy': 0.16, 'Angry':\\ 0.36, 'Surprise': 0.16, 'Sad':\\ 0.17, 'Fear': 0.34\}\end{tabular} \\ \hline
… &
  … &
  … &
  … &
  … &
  … \\ \hline
12 &
  Me before you &
  \begin{tabular}[c]{@{}l@{}}\{'Happy': 0, 'Angry':\\ 0.06, 'Surprise': 0.12,\\ 'Sad': 0.62, 'Fear': 0.19\}\end{tabular} &
  \begin{tabular}[c]{@{}l@{}}\{'Happy': 0.67, 'Angry': 0.33,\\ 'Surprise': 0.0, 'Sad': 0.0,\\ 'Fear': 0.0\}\end{tabular} &
  \begin{tabular}[c]{@{}l@{}}\{'Happy': 0.67, 'Angry':\\ 0.63, 'Surprise': 0.44,\\ 'Sad': 0.56, 'Fear': 0.67\}\end{tabular} &
  \begin{tabular}[c]{@{}l@{}}\{'Happy': 0.34, 'Angry':\\ 0.25, 'Surprise': 0.13, 'Sad':\\ 0.4, 'Fear': 0.21\}\end{tabular} \\ \hline
\end{tabular}}
\end{adjustbox}
\label{predicted_scores}
\end{table*}

\begin{table*}[t]
\centering
\caption{Emotion score of participants’ favorite movies}
\begin{adjustbox}{center}
\resizebox{\linewidth}{!}{%
\begin{tabular}{|l|l|l|l|l|l|}
\hline
id &
  Movie &
  \begin{tabular}[c]{@{}l@{}}Emotion score of movies\\ from description text\end{tabular} &
  \begin{tabular}[c]{@{}l@{}}Emotion score of movies\\ from soundtrack\end{tabular} &
  \begin{tabular}[c]{@{}l@{}}Emotion score of movies\\ from poster\end{tabular} &
  \begin{tabular}[c]{@{}l@{}}Average emotion score\\ of movies\end{tabular} \\ \hline
1 &
  The Notebook &
  \begin{tabular}[c]{@{}l@{}}\{'Happy': 0.45, 'Angry':\\ 0.05, 'Surprise': 0.15,\\ 'Sad': 0.25, 'Fear': 0.1\}\end{tabular} &
  \begin{tabular}[c]{@{}l@{}}\{'Happy': 0.25, 'Angry':\\ 0.0, 'Surprise': 0, 'Sad':\\ 0.5, 'Fear': 0.25\}\end{tabular} &
  \begin{tabular}[c]{@{}l@{}}\{'Happy': 0.56, 'Angry':\\ 0.33, 'Surprise': 0.71,\\ 'Sad': 0.44, 'Fear': 0.56\}\end{tabular} &
  \begin{tabular}[c]{@{}l@{}}\{'Happy': 0.4, 'Angry': 0.08,\\ 'Surprise': 0.19, 'Sad': 0.37,\\ 'Fear': 0.23\}\end{tabular} \\ \hline
2 &
  Split &
  \begin{tabular}[c]{@{}l@{}}\{'Happy': 0.0, 'Angry':\\ 0.22, 'Surprise': 0.11, 'Sad':\\ 0.22, 'Fear': 0.44\}\end{tabular} &
  \begin{tabular}[c]{@{}l@{}}\{'Happy': 0.5, 'Angry':\\ 0.25, 'Surprise': 0, 'Sad':\\ 0.25, 'Fear': 0.0\}\end{tabular} &
  \begin{tabular}[c]{@{}l@{}}\{'Happy': 0.56, 'Angry':\\ 0.5, 'Surprise': 0.33, 'Sad':\\ 0.44, 'Fear': 0.56\}\end{tabular} &
  \begin{tabular}[c]{@{}l@{}}\{'Happy': 0.26, 'Angry':\\ 0.28, 'Surprise': 0.11, 'Sad':\\ 0.27, 'Fear': 0.31\}\end{tabular} \\ \hline
3 &
  Oppenheimer &
  \begin{tabular}[c]{@{}l@{}}\{'Happy': 0.25, 'Angry':\\ 0.0, 'Surprise': 0.25, 'Sad':\\ 0.0, 'Fear': 0.5\}\end{tabular} &
  \begin{tabular}[c]{@{}l@{}}\{'Happy': 0.0, 'Angry': 1.0,\\ 'Surprise': 0, 'Sad': 0.0,\\ 'Fear': 0.0\}\end{tabular} &
  \begin{tabular}[c]{@{}l@{}}\{'Happy': 0.33, 'Angry':\\ 0.43, 'Surprise': 0.25,\\ 'Sad': 0.38, 'Fear': 0.33\}\end{tabular} &
  \begin{tabular}[c]{@{}l@{}}\{'Happy': 0.18, 'Angry':\\ 0.41, 'Surprise': 0.17, 'Sad':\\ 0.06, 'Fear': 0.31\}\end{tabular} \\ \hline
4 &
  Barbie &
  \begin{tabular}[c]{@{}l@{}}\{'Happy': 0.06, 'Angry':\\ 0.03, 'Surprise': 0.09,\\ 'Sad': 0.41, 'Fear': 0.41\}\end{tabular} &
  \begin{tabular}[c]{@{}l@{}}\{'Happy': 0.0, 'Angry': 0.0,\\ 'Surprise': 1, 'Sad': 0.0,\\ 'Fear': 0.0\}\end{tabular} &
  \begin{tabular}[c]{@{}l@{}}\{'Happy': 0.36, 'Angry':\\ 0.3, 'Surprise': 0.3, 'Sad':\\ 0.27, 'Fear': 0.36\}\end{tabular} &
  \begin{tabular}[c]{@{}l@{}}\{'Happy': 0.09, 'Angry':\\ 0.07, 'Surprise': 0.43, 'Sad':\\ 0.25, 'Fear': 0.27\}\end{tabular} \\ \hline
\end{tabular}}
\end{adjustbox}
\label{predicted_fav}
\end{table*}

\subsection{Group Movie Recommendation Example}
Let us show the example of selecting a movie for a group of people using the proposed approach. Given four movie viewers and a pool of 12 movie options (shown in Table \ref{input_12_movies}), we aim to ensure satisfaction and provide the best movie recommendation for the participants.

Each participant provided information about their best-loved movie, a reference point for their preferences (see Table \ref{input_4_movies}). Using the methods described in Section III, we conducted an emotional analysis of three primary sources associated with each movie: the movie poster (picture), the main soundtrack (music), and the movie description (text). Emotional data containing - \textit{Happiness, Anger, Surprise, Sadness,} and \textit{Fear} from the input are shown in Table \ref{predicted_scores}, and emotional data of best-loved movies of each participant is provided in Table \ref{predicted_fav}.



Emotional scores from the three sources were aggregated using \eqref{agg} for each of the 12 offered movies and four favorite movies. After that, we calculated the Jaccard similarity coefficient between each movie's emotional composition and each participant's best-loved movie. This process gives a Jaccard value indicating the similarity between the emotional composition of the 12 movies and the participants' preferred choices, shown in Tables \ref{predicted_scores} and \ref{predicted_fav}. The Box plot of predicted emotion scores is shown in Fig. \ref{boxplot}.

%

The set of movies with the mean Jaccard value equal to the highest coefficient, having the greatest similarity to the participants' best-loved movies, was identified as the best choice for the group. After that, we filtered them by genre. As a result, we obtained two movies with the highest recommendation scores of 0.8, namely “Titanic” and “Me Before You.” The least recommended movie for this group is “Passengers,” with a score of 0.34.


\begin{table}[ht]
\centering
\caption{Feedback from Participants on Movie Recommendation}
    \begin{tabular}{|l|r|r|r|}
    \hline
    \multicolumn{1}{|c|}{\textbf{Participant}} & \multicolumn{1}{c|}{\textbf{\begin{tabular}[c]{@{}c@{}}Agreement score \\ (0-10)\end{tabular}}} & \multicolumn{1}{c|}{\textbf{\begin{tabular}[c]{@{}c@{}}Confidence Level \\ (0-10)\end{tabular}}} & \multicolumn{1}{c|}{\textbf{\begin{tabular}[c]{@{}c@{}}Feedback \\ value\end{tabular}}} \\ \hline
    Participant 1                              & 6                                                                                         & 4                                                                                                & 5.0                                                                                       \\ \hline
    Participant 2                              & 9                                                                                         & 6                                                                                                & 8.44                                                                                    \\ \hline
    Participant 3                              & 5                                                                                         & 5                                                                                                & 4.99                                                                                    \\ \hline
    Participant 4                              & 3                                                                                         & 7                                                                                                & 3.75                                                                                    \\ \hline
    \end{tabular}
\label{tab:participant_feedback}
\end{table}

\begin{figure}[h]
    \centering
    \includegraphics[width=0.5\textwidth]{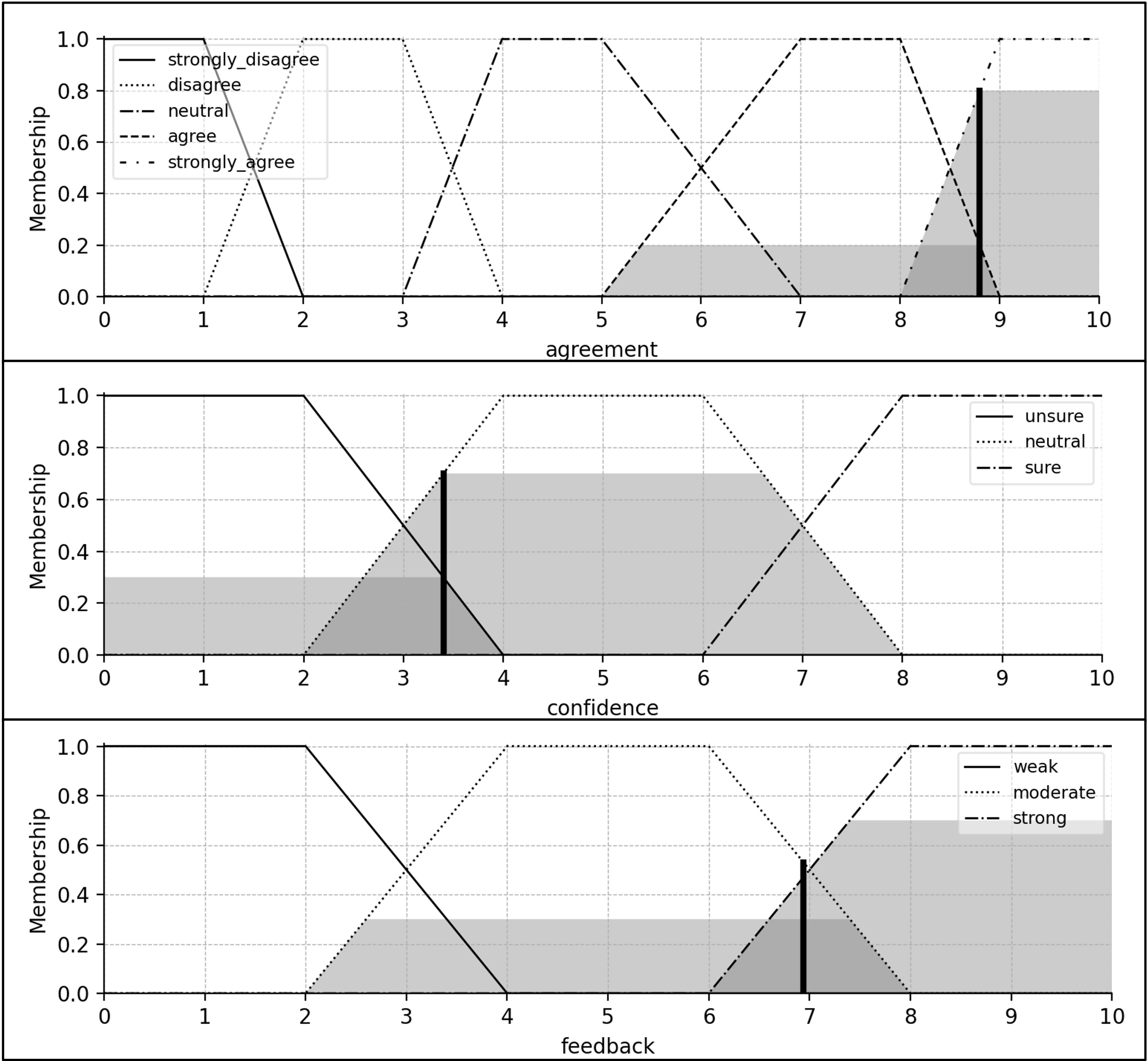}
    \caption{Visual representation of the inference process. For example, the \textit{Agreement} \textit{value} is 8.8, and the\textit{ Confidence} \textit{value} is 3.4.
    As a result, we get a \textit{Feedback value} $\approx$ 6.94.}
    \label{Fuzzy_rules}
\end{figure}

After determining the most recommended movie for the group, the level of consensus with the proposed movie was calculated using fuzzy logic by using participants' feedback. Each participant was asked to rate their agreement with the recommended movie and their confidence levels in these agreements on a scale of 0 to 10, shown in Table~\ref{tab:participant_feedback}. 
The feedback value was calculated based on fuzzy logic rules that used both the agreement and confidence scores, providing a nuanced measure of each participant's satisfaction with the movie recommendation.

For example, let us find the feedback value in the following scenario: the agreement value is 6, and the confidence value is 4. The next step is fuzzification, mapping the crisp numerical values into fuzzy linguistic terms. During fuzzification, each crisp input value is evaluated against these membership functions to determine its degrees of membership in the corresponding fuzzy sets. For instance, an agreement score of 6 has degrees of membership, such as 0.5 in "Agree" and 0.5 in "Neutral". The next important step in a fuzzy inference system is the fuzzy rules. A fuzzy logic system comprises all the rules needed to cover the possible combinations of input linguistic terms. For instance, some rules in our system include:
\begin{itemize}
    \item Rule \#5: IF Agreement is 'Agree' AND Confidence is 'Neutral', THEN Feedback is 'Moderate'
    \item Rule \#7: IF Agreement is 'Neutral' AND Confidence is 'Unsure' THEN Feedback value is 'Moderate'
\end{itemize}
Following the defuzzification process, the centroid method obtained a clear answer. As a result of performing aggregation based on fuzzy rules, we get a feedback value equal to 5. The visual representation of the Rules as an example is shown in Fig. \ref{Fuzzy_rules}. 
The feedback measure of all Participants was calculated and was shown in Table~\ref{tab:participant_feedback}.

To evaluate the consensus among the group members regarding the recommended movie, we calculated the Interquartile Range (IQR) using Equation \ref{iqr} and the mean of the feedback scores. The calculated IQR for the group's feedback scores was $\approx$1.18, and the mean was equal to $\approx$5.54. These values indicate that the participants' satisfaction level dispersion is relatively low. At the same time, the mean value is relatively high, suggesting a generally positive consensus about the movie recommendation, shown in Table \ref{consensus_scores}. The low IQR value and high mean value demonstrated agreement among the group members and approved that the selected movie largely met the collective expectations and preferences of the participants.

\begin{table}[]
\centering
    \caption{Consensus scores of participants}
    \begin{tabular}{|c|l|l|}
    \hline
    \textbf{\begin{tabular}[c]{@{}c@{}}Interquartile Range \\ (IQR)\end{tabular}} & \multicolumn{1}{c|}{\textbf{Mean}} & \textbf{Consensus level}  \\ \hline
    1.18                                                                          & 5.54                               & \multicolumn{1}{c|}{High} \\ \hline
    \end{tabular}
    \label{consensus_scores}
\end{table}

\subsection{Experiment: Emotion Analysis of top 100 movies}
We also conduct experiments to explore the relationship between detected emotions and movie popularity. This research contributes socially by providing insights into the emotional aspects that influence the ratings of movies. Extracting emotions such as \textit{Happy, Angry, Sad, Surprised,} and \textit{Fear} from multiple channels, including posters, music, and movie descriptions, offers a comprehensive understanding of the emotional landscape of movies. Through correlation analysis between these emotional cues and movie ratings using dataset information, this study sheds light on the emotional factors contributing to a movie's perceived quality.

\begin{figure}[tb]
    \centering
    \includegraphics[width=0.45\textwidth]{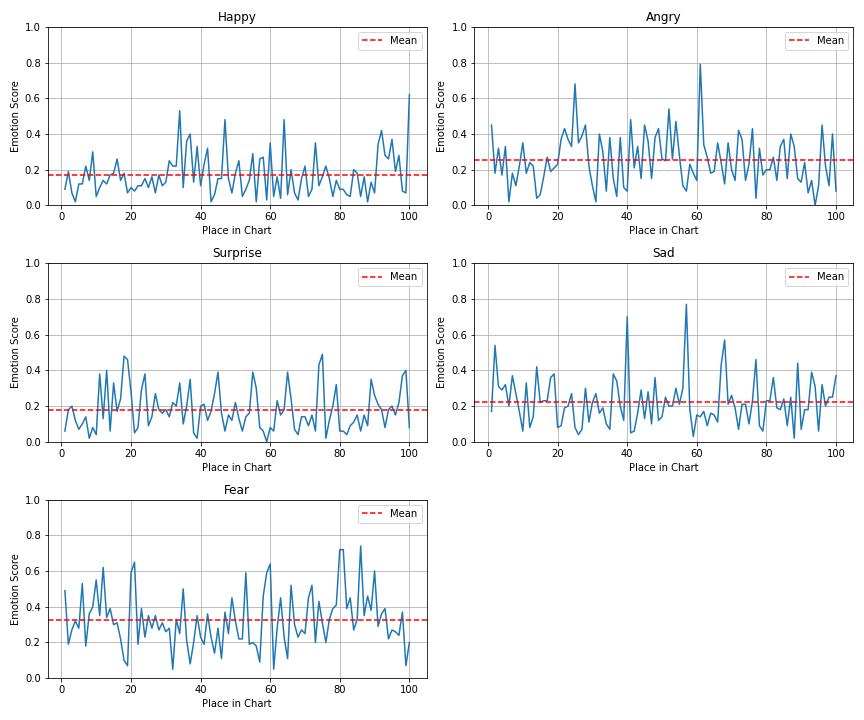}
    \caption{Emotion distribution in 100 most popular movies}
    \label{top100_emotion}
\end{figure}

For this purpose, we use the popular movies chart information from TMDB, which constantly refreshes. We chose and analyzed 100 popular movies. A noteworthy observation is that the data shows a continuous record of emotional variability through the places in rating, with some emotions being more intense than others.
The variations of the emotions depending on the place in the rating are shown in Fig. \ref{top100_emotion}. We can see the differences through ratings by monitoring each emotion line separately. We can observe that in the most popular movies, \textit{Happiness} (approximately the first 30 movies) and \textit{Surprise} (around the first ten movies), scores are much lower than the other partitions. At the same time, in the rest of the graph, they fluctuate intensely. Both of them have a mean value of less than 0.2. This implies that positive feelings are not very common in the emotional landscape of popular movies.
In contrast, \textit{Anger} and \textit{Sadness} exhibit hesitation across the entire graph, many peaks, and a mean value higher than positive emotions. This suggests that the emotional terrain of popular movies tends to be dominated by negative emotions.
But despite all of this variety, \textit{Fear} emerges as the most noticeable feeling. The mean scores for \textit{Fear} have been consistently higher throughout the line plot, indicating that \textit{Horror} themes appear to touch deeply with movies.

\begin{figure}[tb]
    \centering
    \includegraphics[width=0.45\textwidth]{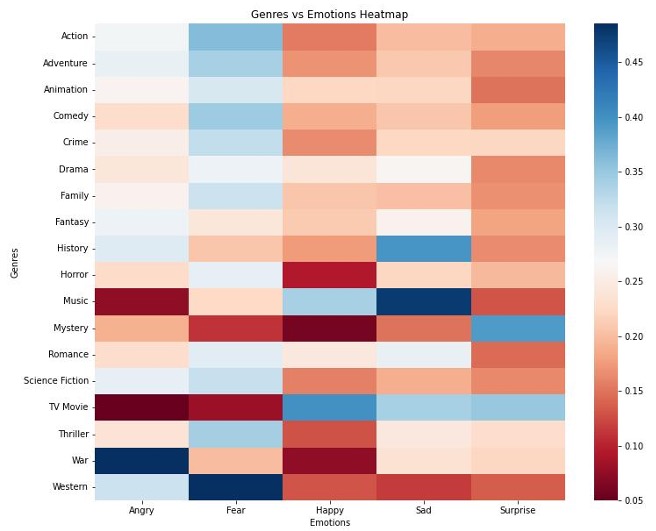}
    \caption{Genre/emotion correlation in 100 most popular movies}
    \label{heatmap}
\end{figure}

\begin{figure}[tb]
    \centering
    \includegraphics[width=0.3\textwidth]{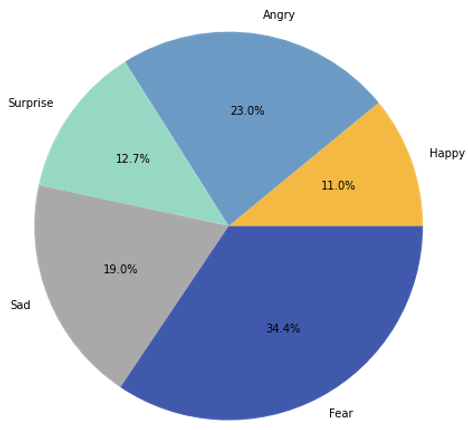}
    \caption{Emotion distribution in 100 most popular movies with threshold 0.2}
    \label{piechart}
\end{figure}

We also explore the coherence between the genre of the movie and its emotional aspect, as illustrated in Fig. \ref{heatmap}. In this graph, we can also see that positive emotions, such as \textit{Happiness} and \textit{Surprise}, are rarely found in most genres; moreover, in the case of existence, the emotional score is much lower. \textit{Sadness} can be considered an emotion with a neutral score, having a middle point in about half of the genres, and taking high values only in \textit{Drama, Fantasy, Romance, History,} and \textit{Music}. \textit{Anger} strongly meets in \textit{War}, and slightly occurs in almost all genres except \textit{Music} and \textit{TV movies}. \textit{Fear} has a strong correlation with \textit{Western} movies and a weak connection with \textit{Mystery} and \textit{TV movies}. Interesting investigations can be observed in the correlation map, most positive genres like \textit{Family, Fantasy,} and \textit{Comedy} strongly correlate with negative feelings rather than positive ones. The overall Emotion distribution in 100 popular movies with a threshold of 0.2 is presented in Fig. \ref{piechart}



\begin{figure}[tb]
    \centering
    \includegraphics[width=0.4\textwidth]{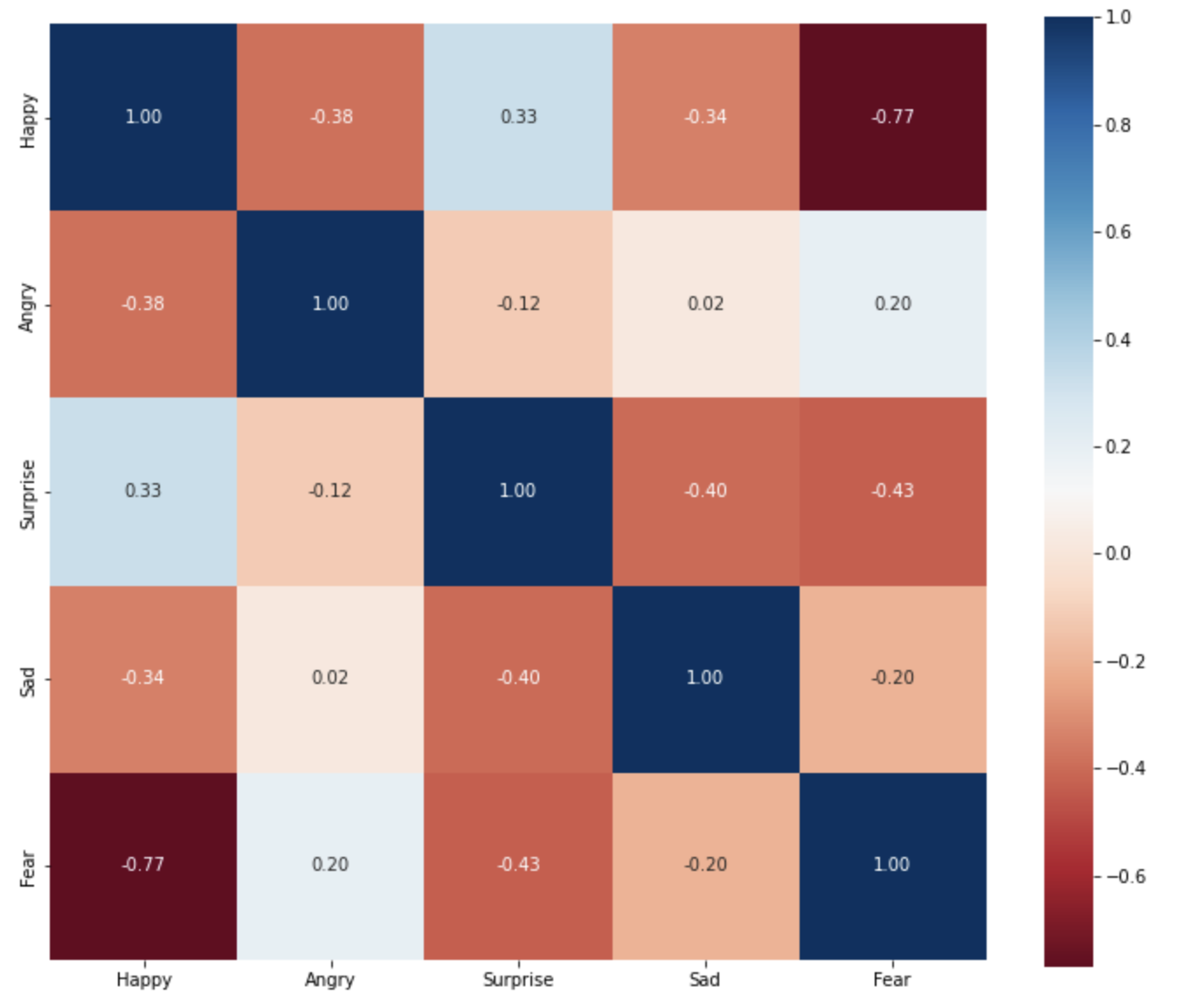}
    \caption{Emotion correlation in survey responses}
    \label{userheatmap}
\end{figure}

We also analyzed correlations between emotions selected by humans in our survey (see Fig. \ref{userheatmap}. As we can see, \textit{Happy} and \textit{Surprise} ($r$=0.33),\textit{ Angry} and \textit{Fear} ($r$=0.20) emotions are highly correlated. In contrast, there is a negative correlation identified between \textit{Happy} and \textit{Fear} ($r$=-0.77), \textit{Happy} and \textit{Angry} ($r$=-0.38), \textit{Surprise} and \textit{Fear} ($r$=0.43) emotions. 

\section{Discussion}
Let us consider how the current study's findings compare to previous studies. Several recent works investigated the influence of emotion in providing movie recommendations \cite{ieee5}, \cite{6}, \cite{pol1}. Some studies \cite{MindFrame}, \cite{MoodIndict}, \cite{movRec2023} proposed movie recommendation systems focusing on providing recommendations based on the emotions detected from user's facial expressions.

Our research findings support previous studies \cite{disc_em}, which have identified significant variations in emotions portrayed in movies, as illustrated by a boxplot. Similar results were obtained by us and are illustrated in Fig. \ref{boxplot}. In addition, our results agree on the prevalence of negative emotions in movies (\textit{Fear, Sad, Angry}) over positive ones. In contrast, our findings presented in Fig. \ref{piechart} do not support the recent research focusing on analyzing emotion dynamics in movie dialogues \cite{disc_em2}, according to which positive word usage in movies is much higher than negative ones.

Recent research identified a high positive correlation between positive emotions collected through EEG present in movies, as opposed to negative emotions \cite{ieeecs4}. Our research findings in Fig.\ref{userheatmap} support this phenomenon, as the correlation scores among positive emotions are higher than negative ones.

 Similar research has been conducted previously but focused on single-user emotion recognition. However, a limited number of works focused on emotion-based recommendations for groups. Most emotion detection methods are still limited to a single channel at the moment \cite{KOZLOV2024771}. Another advantage of our approach lies in multi-channel emotion detection, with subsequent consensus estimation.




\section{Conclusion}


This paper proposes a novel approach to group movie recommendations using multimodal emotion recognition. We examine emotions from multiple sources, including movie descriptions, soundtracks, and posters, providing a comprehensive understanding of the emotions associated with each movie. Then we employ the Jaccard similarity index to match each participant's emotional preferences to prospective movie choices, followed by a fuzzy inference technique to evaluate the group consensus. Our findings suggest that emotion analysis is a promising approach to facilitating consensus in group movie recommendation systems. 


The proposed approach has certain limitations, like a relatively straightforward consensus-building process, which may not fully account for the complexities of human interaction and decision-making in group settings. 
Another limitation is that emotions are inherently complex and may not always be accurately captured through the analysis of text, music, and images alone.

In future work, we plan to collect bigger datasets and include additional types of media, such as movie trailers, actor expressions, and viewer reviews, which could provide a more rounded view of movies' emotional impact. We also plan to integrate traditional recommendation factors like past viewing history with emotional analysis to enhance recommendation accuracy.

\bibliographystyle{IEEEtran}
\bibliography{export}
\begin{IEEEbiography}[{\includegraphics[width=1in,height=1.25in,clip,keepaspectratio]{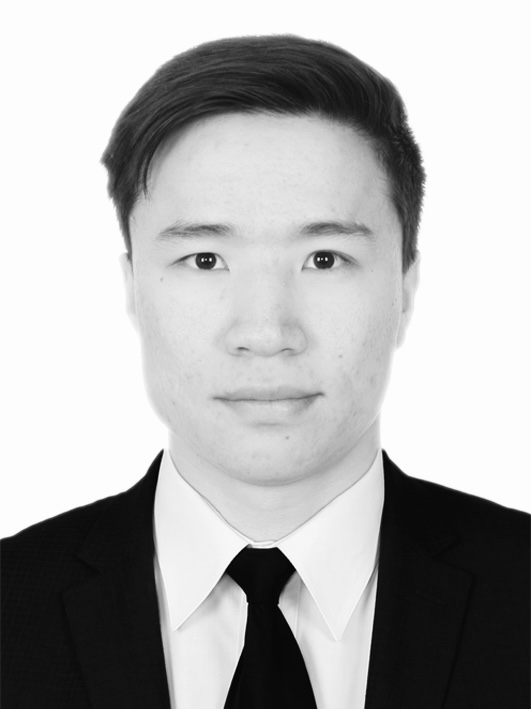}}]{Adilet Yerkin} received a B.S. degree in Information systems from the International University of Information Technologies, Almaty, Kazakhstan, in 2022.  He is currently pursuing an M.S. degree in Data Science with the School of Information Technology and Engineering, at the Kazakh-British Technical University, Almaty. He participated in conferences, such as KBTU AGSRW 2023, IEEE AITU: Digital Generation 2024 and IITU YDF-2024, which received awards for the best paper. Also, he works as a senior data scientist in a leading state IT company in Kazakhstan. His research interests include machine learning, group decision making systems, fuzzy logic and sets.

\end{IEEEbiography}

\begin{IEEEbiography}[{\includegraphics[width=1in,height=1.25in,clip,keepaspectratio]{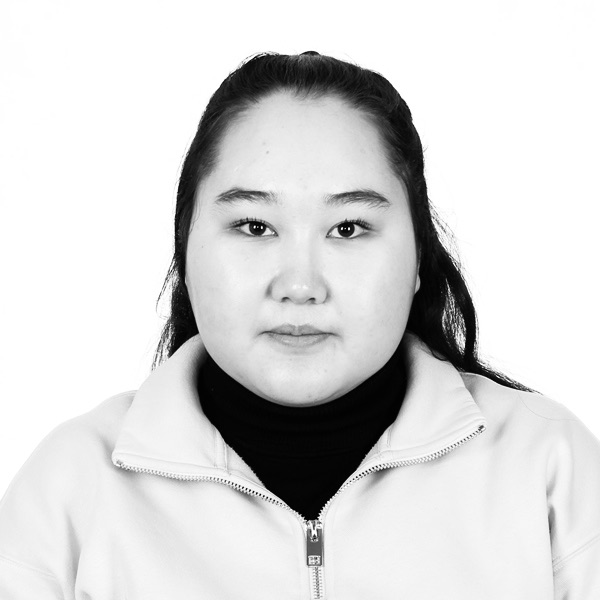}}]{Elnara Kadyrgali} received a B.S. degree in Computer System and Software from
the Kazakh-British Technical University, Almaty, Kazakhstan, in 2022.  She is currently pursuing an M.S. degree in IT management at the same university. She participated in a number of conferences, such as KBTU AGSRW 2023, IITU YDF-2024 and 2024 IEEE AITU: Digital Generation, which received the best paper award. Her research interests include music emotion recognition and recommendation systems.

\end{IEEEbiography}
\begin{IEEEbiography}[{\includegraphics[width=1in,height=1.25in,clip,keepaspectratio]{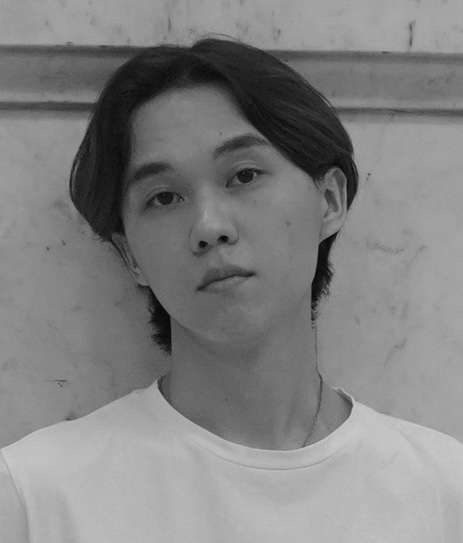}}]{Yerdauit Torekhan} is currently pursuing the bachelor’s degree in information systems with the School of Information Technology and Engineering at the Kazakh-British Technical University, Almaty, Kazakhstan and works as a data engineer in a leading IT company in Kazakhstan. He participated in a conference, 2024 IEEE AITU: Digital Generation, which received the best paper award. His research interests include machine learning, image processing, emotion detection, and human-friendly systems.

\end{IEEEbiography}

\begin{IEEEbiography}[{\includegraphics[width=1in,height=1.25in,clip,keepaspectratio]{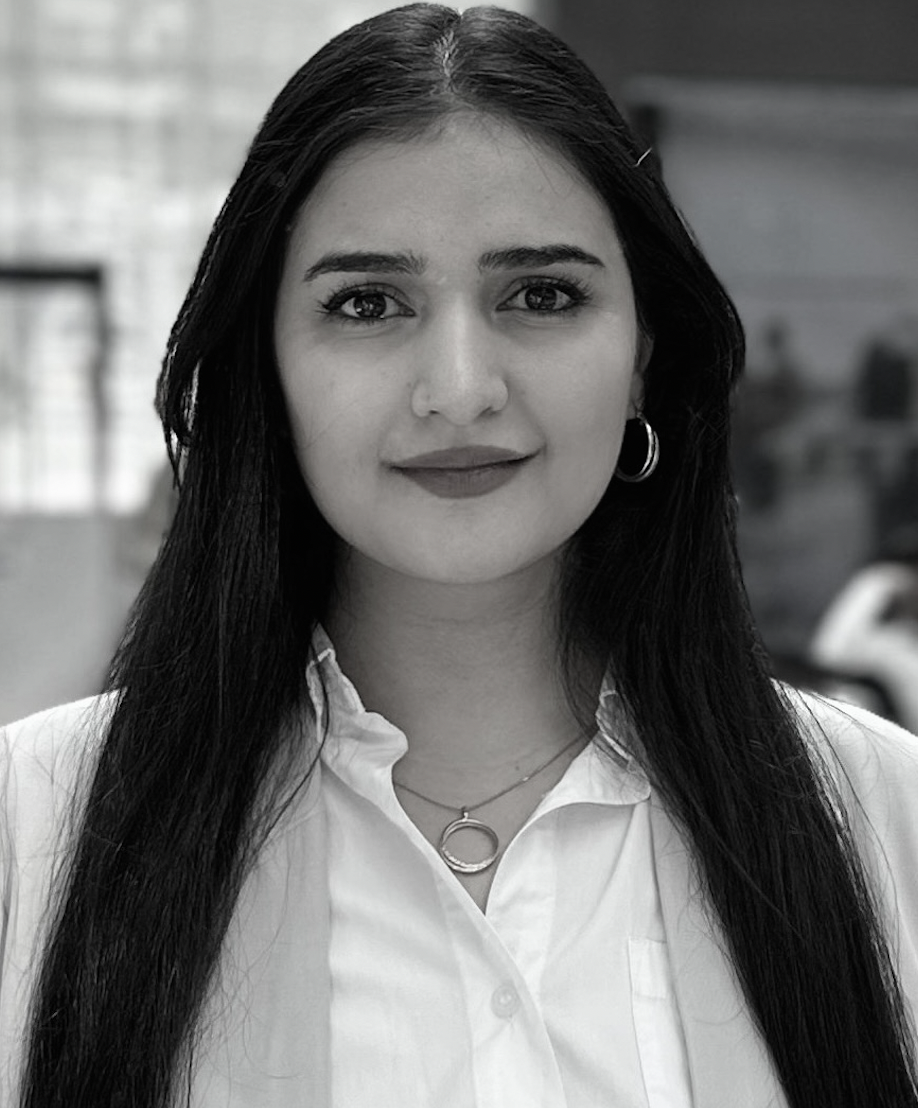}}]{Pakizar Shamoi}  received the B.S. and M.S. degrees in information systems from the Kazakh-British Technical University, Almaty, Kazakhstan in 2011 and 2013, and the Ph.D. degree in engineering from Mie University, Tsu, Japan, in 2019. In her academic journey, she has held various teaching and research positions at Kazakh-British Technical University, where she has been serving as a professor in the School of Information Technology and Engineering since August 2020. She is the author of 1 book and more than 33 scientific publications. Awards for the best paper at conferences were received five times. Her research interests include artificial intelligence and machine learning in general, focusing on fuzzy sets and logic, soft computing, representing and processing colors in computer systems, natural language processing, computational aesthetics, and human-friendly computing and systems. She took part in the organization and worked in the org. committee (as head of the session and responsible for special sessions) of several international conferences - IFSA-SCIS 2017, Otsu, Japan; SCIS-ISIS 2022, Mie, Japan; EUSPN 2023, Almaty, Kazakhstan. She served as a reviewer at several international conferences, including IEEE:
SIST 2023 and 2024, SMC 2022, SCIS-ISIS 2022, SMC 2020, ICIEV-IVPR 2019, ICIEV-IVPR 2018.

\end{IEEEbiography}
\end{document}